\documentclass[times,12pt,preprint]{elsarticle}

\usepackage{amssymb}

\usepackage{lineno,hyperref}
\usepackage{graphicx}
\usepackage{amsmath}
\usepackage{mathrsfs}
\usepackage{algorithm}
\usepackage{algpseudocode}
\usepackage{booktabs,tabularx}
\usepackage{array}
\usepackage{placeins}
\usepackage[font=small,labelfont=bf]{caption}

\usepackage{booktabs}
\usepackage{longtable}
\usepackage{array}
\usepackage{multirow}
\usepackage{wrapfig}
\usepackage{float}
\usepackage{colortbl}
\usepackage{pdflscape}
\usepackage{tabu}
\usepackage{threeparttable}
\usepackage{threeparttablex}
\usepackage[normalem]{ulem}
\usepackage{makecell}
\usepackage{xcolor}

\newcolumntype{C}[1]{>{\centering\let\newline\\\arraybackslash\hspace{0pt}}m{#1}}

\modulolinenumbers[5]

\begin{document}

\begin{frontmatter}



\title{Multi-View Conformal Learning for Heterogeneous Sensor Fusion}


\author[ma]{Enrique Garcia-Ceja\fnref{myfootnote}}

\fntext[myfootnote]{enrique.gc@tec.mx}

\affiliation[ma]{organization={Tecnlogico de Monterrey},
            addressline={Av. Eugenio Garza Sada 2501 Sur}, 
            city={Monterrey},
            postcode={64849}, 
            state={Nuevo Leon},
            country={Mexico}}

\begin{abstract}
Being able to assess the confidence of individual predictions in machine learning models is crucial for decision making scenarios. Specially, in critical applications such as medical diagnosis, security, and unmanned vehicles, to name a few. In the last years, complex predictive models have had great success in solving hard tasks and new methods are being proposed every day. While the majority of new developments in machine learning models focus on improving the \emph{overall} performance, less effort is put on assessing the trustworthiness of \emph{individual} predictions, and even to a lesser extent, in the context of sensor fusion. To this end, we build and test multi-view and single-view conformal models for heterogeneous sensor fusion. Our models provide theoretical marginal confidence guarantees since they are based on the conformal prediction framework. We also propose a multi-view semi-conformal model based on sets intersection. Through comprehensive experimentation, we show that multi-view models perform better than single-view models not only in terms of accuracy-based performance metrics (as it has already been shown in several previous works) but also in conformal measures that provide uncertainty estimation. Our results also showed that multi-view models generate prediction sets with less uncertainty compared to single-view models.
\end{abstract}




\end{frontmatter}


\section{Introduction}
\label{sec:intro}

The prevalence of machine learning models trained with sensor data has been on the rise. This is due to new enabling technologies in wearable devices, Internet of Things, Industry 4.0, and so on. This comes with new challenges, being one of the most important ones, \emph{sensor fusion}. That is, how to aggregate the data from different sensors with the objective of optimizing a given metric. Typically, this metric is the prediction performance of a machine learning model. A system may be composed of several sensors of the same type. For example, an array of accelerometers. On the other hand, systems may also be composed of heterogeneous types of sensors, for example, vision-based (cameras), audio-based (microphones, sonars), movement-based (gyroscopes, accelerometers), etc. This opens up the question of how one can effectively combine heterogeneous types of sensors to build better models. This challenge has been addressed within the context of different applications including emotion recognition~\cite{hosseini2024multimodal,ezzameli2023emotion}, activity recognition~\cite{shaikh2024multimodal, islam2023multi}, and mental health monitoring~\cite{fang2023multimodal, garcia2018mental}, to name a few.

Several approaches to combine sensors have been proposed~\cite{tang2023comparative,garcia2018multi} yielding superior results compared to using individual sensors. One of such approaches is \emph{multi-view} learning~\cite{xu2013survey}. The main idea of multi-view learning is that each observation can be characterized by several feature sets (views). For example, a webpage can be represented with two views. One view can be extracted from the page's text and another view can be extracted from the text of the hyperlinks pointing to that page. In the case of sensor fusion, each sensor can be treated as a different view~\cite{garcia2018multi} which is the approach that we follow in this work.

The majority of research in this area however, focuses on improving the models' \emph{overall} performance in terms of accuracy-based metrics such as precision, recall, F1-Score, and so on, but paying little attention on quantifying the trustworthiness of \emph{individual} predictions. Quantifying the confidence of single predictions is of paramount importance in fields like medicine, finance, security, etc. Providing a measure of confidence in the predictions is crucial for decision making scenarios. Unfortunately, many machine learning models do not have the capability to provide this type of information since they are trained to reduce the overall error without paying attention to individual predictions. Most machine learning classification models generate prediction scores, for example, the softmax function of a neural network. However, those scores are not calibrated and do not represent true probabilities~\cite{guo2017calibration}. While some calibration methods have been developed \citep{karandikar2021soft,mukhoti2020calibrating}, the majority are specific for neural networks, rendering them not portable. On the other hand, Vovk et al. developed a general framework called \emph{conformal prediction} that works with any type of model since it can be implemented as a wrapper \cite{vovk2005algorithmic}. This framework allows any underlying model (classifier or regressor) to generate predictions with individual confidence quantification. In the case of regression, the uncertainty quantification is implemented with confidence intervals. For classification, it is implemented with prediction sets. As opposed to traditional classifiers that generate single-class predictions, conformal models produce prediction sets. Single-class predictions do not offer any uncertainty quantification estimates making it difficult to know when should one trust a given prediction. The single prediction could be completely misleading but a typical classifier is trained to always produce a label as output. That is, they are not able to generate \emph{``I don't know''} answers. Instead of single predictions, for a given input data point, a conformal model will output a prediction set. The prediction set could be empty, meaning that the model is not confident enough to provide an answer. The prediction set can also contain one or more predicted classes (up to $k$, where $k$ is the number of classes). Bigger sets imply more uncertainty, thus, smaller sets are preferred. The predicted set size and the properties of its elements can be used to construct confidence measures (see Section~\ref{sec:measures}).

Given the importance of confidence quantification for individual predictions and its implication in critical applications, in this work we build multi-view conformal models. Our multi-view conformal models have theoretical performance guarantees based on the conformal prediction framework. Specifically, we built two multi-view conformal models and one multi-view semi-conformal model. The first one is based on feature aggregation, the second one is based on multi-view stacking~\cite{garcia2018multi}, and the third one is based on our proposed approach on set intersection. We tested our multi-view conformal and semi-conformal models on two different datasets with heterogeneous sensor settings. We are not aware of any other study that thoroughly analyzes multi-view conformal models in terms of traditional and \emph{conformal measures}. The two hypotheses that we aim to test in this work are the following:

\begin{enumerate}
    \item Multi-view models produce predictions with less uncertainty (smaller prediction sets) compared to single-view models.
    \item Multi-view models perform better than single-view models in terms of conformal prediction measures.
\end{enumerate}

This document is organized as follows. Section~\ref{sec:background} presents the fundamental foundations this work is based on. This includes an introduction to sensor fusion methods, multi-view learning, conformal prediction, and conformal prediction performance measures. In Section~\ref{sec:mvcm} we detail the multi-view models used in our evaluations. Section~\ref{sec:datasets} describes the two used datasets in this study and the feature extraction process. In Section~\ref{sec:experiments} we explain the experiments and present the results. Finally in Section~\ref{sec:conclusions} we present our conclusions and future work.

\section{Background}
\label{sec:background}

In this section we describe the topic of sensor fusion. Then, we present an overall introduction to multi-view learning followed by an introduction to conformal prediction. Finally, we detail the conformal prediction measures used in this work.

\subsection{Sensor Fusion}
\label{sec:sf}

Nowadays sensor fusion has become an important topic given the rise of heterogeneous types of available sensors. In the context of machine learning, sensor fusion relates to the set of algorithms that aim to optimize predictive models by intelligently combining information from  diverse sources~\cite{blasch2021machine}.

In supervised learning settings, there exist two main sensor fusion strategies, feature-level fusion and decision-level fusion~\cite{gravina2017multi}. In feature-level fusion, features extracted from different sensors are aggregated into a single high-dimensional feature vector. The advantage of this approach is that it is easy to implement. On the other hand, it can lead to very high-dimensional feature vectors which can cause loss of precision and increased training times. To alleviate this, feature selection algorithms can be used to find the most important features and discard the rest~\cite{tang2014feature}. In decision-level fusion, the final prediction is obtained by aggregating individual decisions from multiple sensors. It is often the case that the aggregated decision will be of higher quality compared to any of the individual decisions. The individual decisions can be obtained from homogeneous or heterogeneous sensors. A common way of aggregating the individual decisions is by majority voting or by training a meta-learner using the individual decisions as features, as in generalized stacking~\cite{Wolpert1992}.

Sensor fusion techniques have the potential to be applied in many domains. This can help to optimize processes, making them more robust. For example, sensor data fusion has been used for emotion recognition. An example of this, is the work of Cimtay et al.~\cite{cimtay2020cross}. In their work, they proposed a multimodal method based on a feature and decision level fusion convolutional neural network to recognize emotions (sad, neutral, and happy) based on different sources of information including: facial expressions, galvanic skin response, and electroencephalogram (EEG). In the work of de la Fuente et al.~\cite{fuente2023multimodal}, they tested feature-level and decision-level deep learning models to recognize frustration during game-play. They considered audio and video information and their decision-level approach obtained better results. Sensor fusion has also been applied for activity recognition tasks~\cite{qiu2022multi}. For example, physical activities recognition has traditionally been carried out using either inertial~\cite{logacjov2024self,thakur2024permutation,middya2024activity,dang2024assessing,baroudi2024classification} or vision-based sensors~\cite{cob2024new,bhola2024review}. Some other works tried combining both types of sensors~\cite{majumder2020vision,hao2024egocentric}. Inertial and Wi-Fi sensors have also been used in combination for activity recognition\cite{guo2024human,GarciaCeja2012,garcia2014contextualized}. Our work mainly differs from previous ones in that we build and evaluate multi-view conformal vs. single-view conformal models that provide confidence estimates through prediction sets instead of single-label predictions.

\subsection{Multi-View Learning}
\label{sec:mvl}

During a feature engineering process, it is not uncommon to find out that an observation can be represented by two or more types of features. Each, providing a different perspective (view) of the entity under consideration. For instance, a movie can be characterized by the audio, the consecutive images, and subtitles (if present). In this case we would have three views: images, audio, and subtitles. If we wanted to automatically assign movies into genres (action, drama, comedy, etc.) we would need to build a classifier. The input would need to be a set of features extracted from each of the three views. Then, those features can be aggregated into a single feature vector and added to the training set to train a single model. However, this approach may not be optimal since individual views possess different statistical properties~\cite{Xu2013} and contribute with different information. For instance, it would be more suitable to use a specific type of model for each of the views based on their characteristics. For example, a convolutional neural network for the images, a recurrent neural network for the audio, and a neural network with embedding layers for the subtitles. Multi-view learning aims to optimize the way several views are combined with the objective of increasing the final performance of a model.

The co-training algorithm proposed by Blum \& Mitchell~\cite{Blum1998} is one of the earliest forms of multi-view learning. This method was developed to train semi-supervised models~\cite{Chapelle2006}. That is, models that not only use labeled data but also unlabeled data to learn from. It was designed for problems that can be decomposed into two views. For example, webpage classification. In this case, one view can be generated by extracting features from the text within the page and another view can be generated with features extracted from the links pointing to the webpage. The algorithm starts by training two independent models (one per view) using the labeled data. Then, perform $n$ iterations or stop when the unlabeled data is exhausted. In each iteration, one of the models will generate label predictions for a subset of the unlabeled data points. Then, those newly labeled data points are added to the training set of the other model and vice versa. This results in the training sets for each view being augmented. The co-training algorithm has the assumption of conditionally independent views. Zhou \& Li proposed a new method called tri-training which uses three classifiers instead of two and does not rely on the view independence assumption \cite{Zhou2005}. 

More recently, multi-view learning approaches have also been proposed for supervised learning settings. For instance, Zhang et al. proposed a deep multi-view learning method that works when views may have missing information~\cite{zhang2020deep}. Lupión et al. developed a multi-view algorithm for 3D pose estimation based on thermal vision sensors using neutral networks~\cite{lupion20243d}. Piriyajitakonkij et al. also proposed a method based on neural networks for recognizing movement during sleep \cite{piriyajitakonkij2020}. They generated two views by extracting time and frequency domain features. Even though the previous approaches have shown promising results, they are specific for neural network architectures. On the other hand, more general methods have also been developed. For example, Garcia-Ceja et al. proposed a multi-view learning approach based on stack generalization that does not rely on any particular underlying model~\cite{garcia2018multi} (see Section~\ref{sec:stacking}). In multi-view stacking, one must select the underlying models (first-level learners and meta-learner). van Loon et al. conducted experiments to find which is the best meta-learner for a specific application (gene-expression)~\cite{van2020view}. In this work we built and tested three multi-view models for sensor data fusion. We consider the data from each sensor as a different view.

\subsection{Conformal Prediction}
\label{sec:cp}

Nowadays, predictive models have become of vital importance in many everyday operations. With the new technological advancements in hardware and computing capabilities, bigger and more complex algorithms are being built. These new complex algorithms can solve problems that some years ago were thought to be only solvable by humans; however they are still designed with the objective of reducing the \emph{overall} average error. However, the quality of single predictions is difficult to assess. Some machine learning models output classification scores (e.g., neural networks' softmax layer), but those scores are not calibrated as noted in ~\cite{guo2017calibration,minderer2021revisiting}. Different calibration techniques for neural networks have been developed~\cite{karandikar2021soft,mukhoti2020calibrating} but are not general enough to be applicable to other types of models. Having access to confidence scores is crucial for applications in medicine, security, finance, and so on. Without this information, it becomes difficult to make decisions with confidence and it is hard to tell if we can trust a particular prediction or not.

\begin{figure}[h!]
\centering
\includegraphics[width=.8\textwidth]{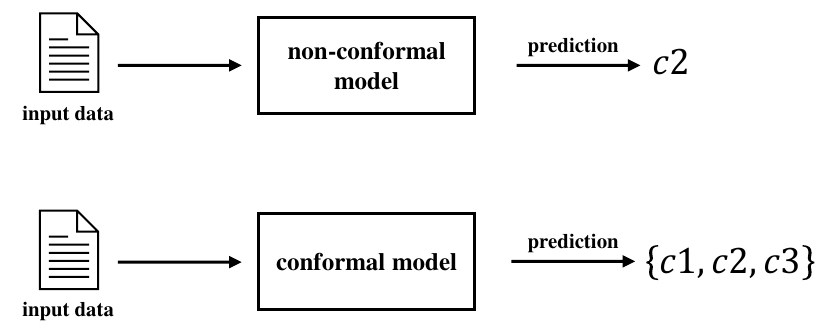}\hfill
\caption{non-conformal vs. conformal model. The conformal model produces prediction sets.}
\label{fig:cp-example}
\end{figure}

To address this, \emph{conformal prediction}, a framework developed by Vovk et al.~\cite{vovk2005algorithmic} can be adapted to almost any classifier or regressor since it works as a wrapper and it produces confidence scores for individual predictions. While non-conformal classifiers produce single-label predictions, conformal models generate prediction sets. Figure~\ref{fig:cp-example} depicts an example. Given the same input data, the non-conformal model generates a single-label prediction while the conformal model produces a \emph{prediction set}. The set could be empty, meaning that the model is not confident enough to make a prediction (an \emph{``I don't know answer''}). In the example, if classes $c1...cn$ represented types of diseases, it would be risky to make a final diagnosis since we are not sure how confident the non-conformal model is with its prediction $c2$ since it could be completely misleading. Even though the prediction set for the conformal model has three possible classes, it is guaranteed that on average and with some probability $p$ the prediction sets contain the true class, thus providing more information for the decision maker.

In order to calculate the final prediction set, we need to specify an underlying classifier $c$ for example, a Random Forest. We also need a \emph{calibration set} which must be independent from the training set but drawn from the same distribution. Finally, we need to define a \emph{non-conformity function} $NC(x,y,c)$ where $x$ is an instance and $y$ is its true label. The non-conformity function captures how strange is the label $y$ for the given instance. A common non-conformity function for classification is $NC(x,y,c) = 1-S(x,y,c)$ where $S$ returns the confidence score as estimated by $c$ for label $y$.

Before starting to make predictions on new query examples, a calibration phase needs to be carried out. This phase consists of computing the non-conformity score $\alpha_i$ for every instance $i$ from a calibration set of size $n$. The score is computed with the non-conformity function that is, $\alpha_i = NC(x_i,y_i,c)$.

When making a prediction for a new example $x_{n+1}$, the p-value $p_{n+1}$ for each label is computed. If the p-value is greater than the predefined error $\epsilon$, the label is included in the prediction set. Thus, a conformal classifier is defined as:

\begin{equation} \label{eq:conformalpredictor}
 \Gamma^{\epsilon} \left( x_1,y_1,\ldots,x_{n},y_{n}, x_{n+1} \right) = y \in Y
\end{equation}

where $y \in Y$ is the set of conformal labels that satisfy:

\begin{equation} \label{eq:pvalue}
 \frac{\lvert \{i=1,\ldots,n:\alpha_i \geq \alpha_{n+1} \} \rvert}{n+1} > \epsilon
\end{equation}

that is, the left hand side of Eq.~\ref{eq:pvalue} computes the p-value for a particular label. By doing so, the probability of including the correct label is no less than $1 - \epsilon$. The maximum tolerated error $\epsilon$ is a parameter specified by the user.

The cardinality of the resulting prediction set is an indication of the confidence. A set size with many elements indicates higher uncertainty. In conformal prediction, small prediction sets are preferred. 

A conformal model is not evaluated by typical performance metrics but based on the quality of the prediction sets. The next section details the conformal prediction measures used in this work.

\subsection{Conformal Prediction Performance Measures}
\label{sec:measures}

The following measures used in conformal prediction are described in detail by Vovk et al. \cite{vovk2016criteria}. These are the measures used in our experiments.

\subsubsection*{Coverage}
The coverage is the percentage of prediction sets where the true label is included:
\begin{equation}
    \frac{1}{k} \sum_{i=1}^k \begin{cases}
      1, & \text{if}\ y_i \in \Gamma^\epsilon_i \\
      0, & \text{otherwise}
    \end{cases}
\end{equation}

where $y$ is the true label.

\subsection*{N criterion (set size)}
The \emph{N criterion} (\emph{setsize}) is the average cardinality of the prediction sets. Smaller values are preferred.
\begin{equation}
    \frac{1}{k} \sum_{i=1}^k |\Gamma^\epsilon_i|
\end{equation}

where $k$ is the number of prediction sets (number of test examples).

\subsubsection*{Percentage of Empty Sets}
The percentage of empty sets (\emph{pctempty}) is:
\begin{equation}
    \frac{1}{k} \sum_{i=1}^k \begin{cases}
      1, & \text{if}\ |\Gamma^\epsilon_i| = 0 \\
      0, & \text{otherwise}
    \end{cases}
\end{equation}

\subsubsection*{M criterion}
The M criterion refers to the percentage of prediction sets that have more than one element. Smaller values are preferred.

\begin{equation}
    \frac{1}{k} \sum_{i=1}^k \begin{cases}
      1, & \text{if}\ |\Gamma^\epsilon_i| > 1 \\
      0, & \text{otherwise}
    \end{cases}
\end{equation}

\subsubsection*{F criterion}
The F criterion refers to the average fuzziness. The fuzziness of an example $x_i$ is the sum of the p-values (excluding the largest one). Smaller values are preferred.

\begin{equation}
    \frac{1}{k} \sum_{i=1}^k \left( \sum_y p^y_i - \underset{y}{\mathrm{max}} p^y_i \right)
\end{equation}

\subsubsection*{Jaccard index}
The \emph{Jaccard index} is used to measure the similarity between sets. A value of $1$ means that the resulting set has one element and is equal to the ground truth $y_i$. Values close to $1$ are preferred:
\begin{equation}
    \frac{1}{k} \sum_{i=1}^k \frac{|\Gamma^\epsilon_i \cap \{y_i\}|}{|\Gamma^\epsilon_i \cup \{y_i\}|}
\end{equation}

\subsection*{OM criterion}
The proportion of prediction sets that include one or more false classes. Smaller values are preferred.

\begin{equation}
    \frac{1}{k} \sum_{i=1}^k \begin{cases}
      1, & \text{if}\ \{\Gamma^\epsilon_i \diagdown \{y_i\} \neq \emptyset\} \\
      0, & \text{otherwise}
    \end{cases}
\end{equation}

\subsection*{OF criterion (observed fuzziness)}
The average of the sum of the p-values of the false labels. Smaller values are preferred.

\begin{equation}
    \frac{1}{k} \sum_{i=1}^k \sum_{y\neq y_i} p^y_i
\end{equation}

\subsection*{OU criterion (observed unconfidence)}
The average unconfidence. The unconfidence of a prediction set is the largest p-value of the false classes. Smaller values are preferred.

\begin{equation}
    \frac{1}{k} \sum_{i=1}^k \underset{y\neq y_i}{\mathrm{max}}\ p^y_i
\end{equation}

\subsection*{OE criterion (observed excess)}
The average number of false classes in the prediction sets. Smaller values are preferred.

\begin{equation}
    \frac{1}{k} \sum_{i=1}^k |\Gamma^\epsilon_i \diagdown \{y_i\}|
\end{equation}

\section{Multi-View Conformal Models}
\label{sec:mvcm}

In this section we explain the three multi-view models used in this work namely, Multi-View Aggregation (MV-A), Multi-View Stacking (MV-S), and Multi-View Intersection (MV-I).

\subsection{Multi-View Aggregation (MV-A)}
\label{sec:aggregation}

\begin{figure}[!h]
\centering
\includegraphics[scale=0.45]{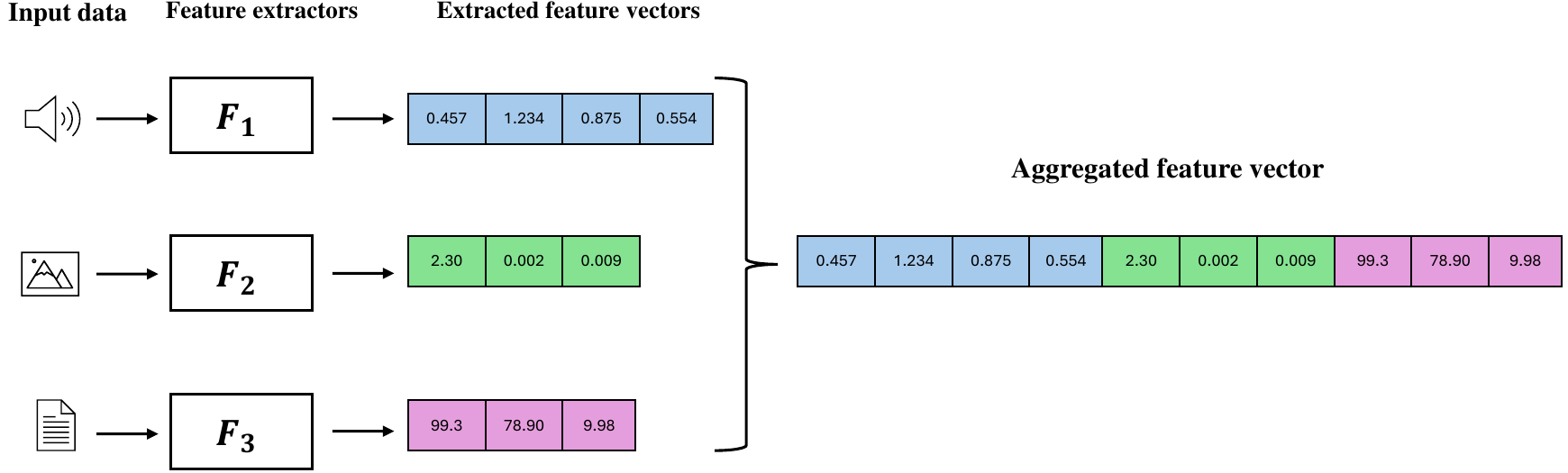}
\caption{Example of aggregating feature vectors from three views.}
\label{fig:aggregation}
\end{figure}

This model is built by aggregating the features of all views into the same feature vector and then, training a single underlying model. Let $\{v_1, v_2, ... v_n\}$ be the set of feature vectors corresponding to $n$ different views. Then, the aggregated feature vector is:

\begin{equation} \label{eq:feature-vector-aggregated}
 v^{*} = v_1 \cdot v_2 \cdot ... \cdot v_n
\end{equation}

where $\cdot$ is the vector concatenation operation. Figure~\ref{fig:aggregation} depicts an example of the procedure of feature vector aggregation for three views.

The aggregated feature vectors $v^{*}_{i}$ of every training point $i$ are used to train the underlying model as usual and then, it is calibrated using the conformal prediction algorithm (Section~\ref{sec:cp}).

\subsection{Multi-View Stacking (MV-S)}
\label{sec:stacking}

The Multi-View Stacking (MV-S) algorithm~\cite{garcia2018multi} is based on stacked generalization proposed by Wolpert (1992)~\cite{Wolpert1992} and is a type of ensemble algorithm that combines the results from several models.

The algorithm first trains a set of models (known as \emph{first-level learners}) using the initial training set. Then, the first-level learners are used to predict the labels of the training set. Those labels are then used as features to train another learner known as the \emph{meta-learner}. The following steps detail the stacking process:

\begin{enumerate}
\item Define a set $\mathscr{L}$ of first-level learners and a \emph{meta-learner}.

\item Train the first-level learners in $\mathscr{L}$ with the training set $\textbf{D}$.  $\textbf{D}$ has $n$ training examples (rows).

\item Predict the classes of $\textbf{D}$ with each of the learners in $\mathscr{L}$. Each learner in $\lvert\mathscr{L}\lvert$ produces a prediction vector $\textbf{p}^i$ of $n$ elements.

\item One-hot encode the prediction vectors $\textbf{p}^i$. Thus, they become matrices $\textbf{P}^i_{n \times k-1}$ where $n$ is the number of instances in the training data $\textbf{D}$ and $k$ is the number of classes. To avoid the dummy variable trap, one of the columns is removed, thus the number of columns becomes $k - 1$.

\item Build a matrix $\textbf{M}_{n \times \lvert\mathscr{L}\lvert (k-1) + k}$  by column binding the prediction one-hot encoded matrices $\textbf{P}^i$ and the averaged class scores $\textbf{S}_{n \times k}$.

\item Build a new training set $\textbf{D}'$ with the matrix $\textbf{M}$ and the true labels $\textbf{y}$.

\item Train the \emph{meta-learner} using $\textbf{D}'$

\item Return the trained stacking model $Model=(\mathscr{L},\textit{meta-learner})$.
\end{enumerate}

\begin{figure}[!h]
\centering
\includegraphics[scale=0.75]{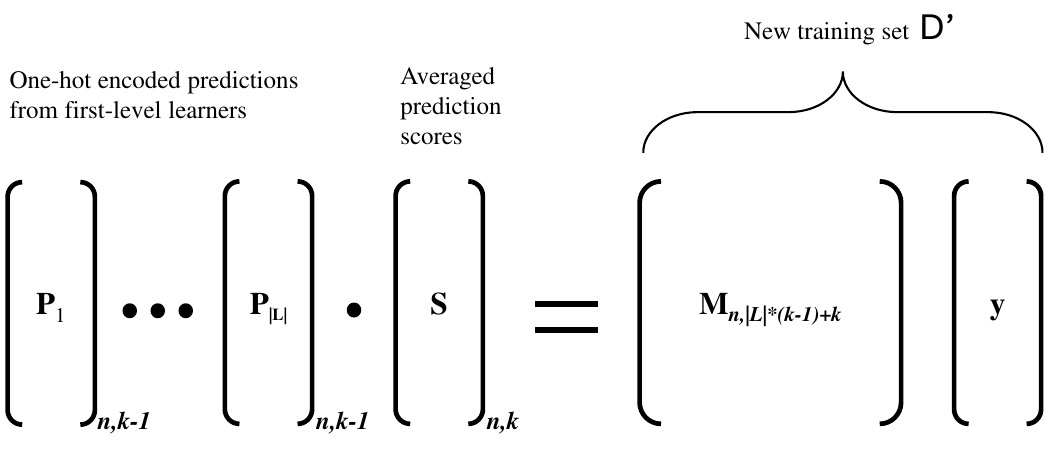}
\caption{The training data $\textbf{D'}$ is constructed by column binding the one-hot encoded label predictions of the first-level learners, the averaged scores, and the true labels \textbf{y}.}
\label{fig:process}
\end{figure}

Figure \ref{fig:process} depicts the steps to build the training data $\textbf{D}'$ that is used to train the meta-learner.

In the second and third step of the algorithm, over-fitting can happen because the predictions are made on the same training set that was used to train the first-level learners. To avoid overfitting, steps $2$ and $3$ can be implemented with $k$-fold cross validation. After $\textbf{D}'$ has been built, the learners in $\mathscr{L}$ are retrained using all the training data in $\textbf{D}$.

It has been shown that adding confidence scores as extra information in matrix $\textbf{M}$ can increase the performance \cite{Ting1999}. Thus, the averaged (across first-level learners) confidence scores for each class are added (matrix $\textbf{S}$ in Figure~\ref{fig:process}).

The MV-S algorithm consists of training one first-level learner for each of the views and aggregating their predictions with the stacked generalization approach. Finally, the multi-view stacking model is calibrated using the calibration data.

\subsection{Multi-View Intersection (MV-I)}
\label{sec:intersection}

In this work, we also propose a Multi-View Intersection (MV-I) model that consists of $n$ first-level conformal models (one model for each of the views). Each first-level learner $l_i$ is independently trained and calibrated with the conformal prediction method (see Section ~\ref{sec:cp}). The prediction set for a new test data point is obtained by computing the intersection of the first-level learners' prediction sets.

\begin{figure}[!h]
\centering
\includegraphics[scale=0.57]{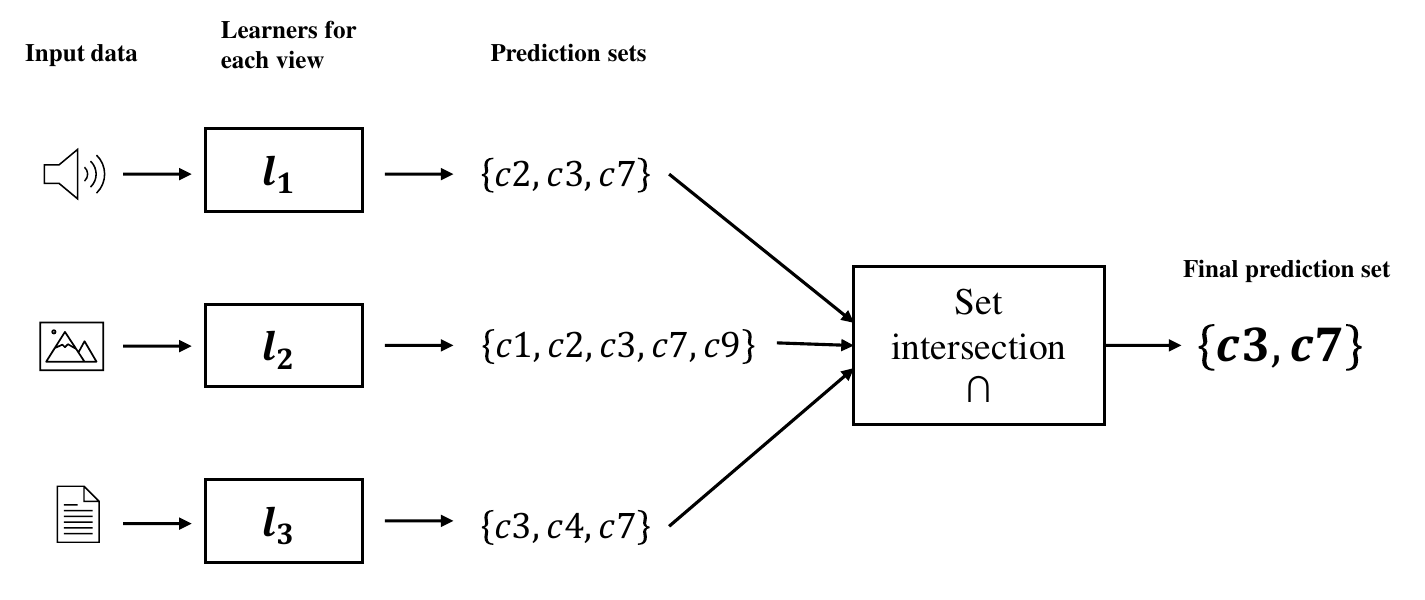}
\caption{Example of MV-I with three views (audio, images, and text).}
\label{fig:intersection}
\end{figure}

The single-label prediction of the non-conformal version of the model is obtained by choosing the label from the intersection set with the highest average score. If the intersection results in an empty set, the predicted label is the one with the highest average score from the union of the predictions sets. Figure~\ref{fig:intersection} shows an example of the MV-I model with three views (audio, images, and text). Each learner produces a prediction set. The final prediction is the intersection of the individual prediction sets. We call this model semi-conformal because the individual learners are conformal but after the intersection operation, the final sets are not necessarily conformal.

\section{Datasets and Feature Extraction}
\label{sec:datasets}

In this section we describe the multi-modal datasets used in our experiments and their extracted features. 

\subsection{HTAD Dataset}
\label{sec:htad}

The Home-Tasks Activities Database (HTAD)~\cite{garcia2021htad} consists of wrist-accelerometer and audio data from individuals conducting $7$ home activities including: mopping floor, sweeping floor, typing on computer keyboard, eating chips, brushing teeth, washing hands, and watching television. The database includes accelerometer and sound features. Three volunteers collected the data. One female and two males with an age range of $25-30$. The volunteers recorded each activity for approximately $3$ minutes. The participants used a wrist-band and a smartphone to collect the data. The accelerometer sampling rate was $31$ Hz and the microphone was $8000$ Hz. This database includes features extracted from $3$-second non-overlapping windows. From each window, two sets of features were extracted. The first one corresponds to the accelerometer sensor and the second one to the sound data. Each set of features corresponds to one of the views (accelerometer and audio).

There are $16$ accelerometer features: The \emph{mean}, \emph{standard deviation}, \emph{max} value for the x, y, and z axes, \emph{pearson correlation} of pairs of axes (xy, xz, and yz), \emph{mean magnitude} (Eq. \ref{eq:magnitude}), \emph{standard deviation of the magnitude}, the \emph{magnitude area under the curve} (AUC, Eq. \ref{eq:auc}) , and \emph{magnitude mean differences} between consecutive readings (Eq. \ref{eq:mean_dif}).

\begin{equation} \label{eq:auc}
AUC = \sum\limits_{t = 1}^T {magnitude(t)}
\end{equation}

\begin{equation} \label{eq:mean_dif}
meandif = \frac{1}{{T - 1}}\sum\limits_{t = 2}^T {magnitude(t) - magnitude(t - 1)}
\end{equation}

\begin{equation} \label{eq:magnitude}
Magnitude(x,y,z,t) = \sqrt {{a_x}{{(t)}^2} + {a_y}{{(t)}^2} + {a_z}{{(t)}^2}}
\end{equation}

From the sound data, $36$ Mel Frequency Cepstral Coefficients (MFCCs) were extracted. The $3$ second sound signals were divided into $1$ second windows. Then, $12$ MFCCs were extracted from each. After feature extraction, $1386$ data points (instances) were generated.

\subsection{Berkeley MHAD Dataset}
\label{sec:berkeley}

This database~\cite{ofli2013} was recorded using data from accelerometers, an optical motion capture system, microphones, depth sensors, and stereo cameras. The data was recorded by $12$ individuals performing $11$ actions. All individuals recorded $5$ repetitions for each action including: 1) jumping in place, 2) jumping jacks, 3) bending, 4) punching, 5) waving two hands, 6) waving one hand, 7) clapping, 8) throwing a ball, 9) sit/stand up, 10) sit down, and 11) stand up.

This resulted in $660$ examples. $2$ recordings were lost because of sensor missing values. In our study we included $3$ views: wrist-acceleration, sound from microphones, and $3$D skeleton points from the motion capture system.

The same features as the HTAD dataset were used for the accelerometer (see Section \ref{sec:htad}) and the same features from the sound data were extracted as detailed in Section \ref{sec:htad}. $12$ MFCCs were extracted from each of the $4$ microphones.

The features from the motion capture device were computed as the distance between an initial joint point (the spine) and the other $30$ joint points. This was done for each frame. Then, the \emph{mean, max} and \emph{min} values from all frames were computed.

\section{Experiments and Results}
\label{sec:experiments}

For the two datasets, we built single-view conformal models for each of the data sources, $2$ multi-view conformal models: Multi-View  Aggregation (MV-A) and Multi-View Stacking (MV-S), and one multi-view semi-conformal model: Multi-View Intersection (MV-I). For the HTAD dataset, two single-view models were built. One with the audio data, and another one with the accelerometer data. For the Berkely-MHAD dataset, three single-view models were built for each of the data sources including audio, accelerometer, and skeleton, respectively. For our experiments, we used Random Forest~\cite{Breiman2001} as the underlying classifier for the single-view models and for MV-A and MV-I. We also used Random Forest for the first-level learners and for the meta-learner for MV-S. The reason is that Random Forest has been shown to produce good overall results~\cite{Fernandez2014} and requires minimal tuning.

For both datasets, $50\%$ of the data was randomly assigned to the training set, $25\%$ to the test set, and $25\%$ was assigned to the calibration set. The confidence was set to $\gamma = 0.95$ (equivalent to an error $\epsilon = 0.05$). To account for variance, the experiments were repeated $15$ times. The LAC method (Least Ambiguous set-valued Classifier) was used as the non-conformity function and is defined as one minus the score of the true label\footnote{MAPIE documentation: \url{https://mapie.readthedocs.io/en/latest/theoretical_description_classification.html}}. The code was implemented in Python using the MAPIE library~\cite{mapie}. The code to reproduce the results is available here: \url{https://github.com/enriquegit/multiview-conformal-prediction-paper}.

Table~\ref{tab:htad_results} shows the results for the HTAD dataset. The first four metrics are the classical ones. The rest of the measures are the conformal ones. An * indicates that smaller values mean better performance. In this case, MV-S was the best model across all measures except for coverage. All models (except MV-I) were very close to the specified $95.0\%$ confidence target. In some cases, it was some decimal units lower. This can be due to some out of distribution samples whose properties were not captured in the calibration set. This is an expected behavior since MV-I is not strictly conformal as discussed in Section~\ref{sec:intersection}. Even though MV-I is not strictly conformal, is was superior than MV-A in almost all measures. For the majority of the non-conformal and conformal measures, the multi-view models performed better compared to the single-view ones.

\begin{table}[h!]
\small
\caption{HTAD average results. * Smaller values are preferred.}
\label{tab:htad_results}
\centering
\begin{tabular}[t]{llllll}
\toprule
  & \textbf{audio} & \textbf{accelerometer} & \textbf{MV-A} & \textbf{MV-S} & \textbf{MV-I}\\
\midrule
\cellcolor{gray!6}{accuracy} & \cellcolor{gray!6}{77.77$\pm$2.06} & \cellcolor{gray!6}{83.75$\pm$2.01} & \cellcolor{gray!6}{88.51$\pm$1.90} & \cellcolor{gray!6}\textbf{90.66$\pm$1.01} & \cellcolor{gray!6}{90.36$\pm$1.66}\\
sensitivity & 76.92$\pm$2.32 & 82.74$\pm$2.05 & 87.73$\pm$2.09 & \textbf{89.96$\pm$1.10} & 89.60$\pm$1.77\\
\cellcolor{gray!6}{specificity} & \cellcolor{gray!6}{96.25$\pm$0.35} & \cellcolor{gray!6}{97.30$\pm$0.34} & \cellcolor{gray!6}{98.09$\pm$0.32} & \cellcolor{gray!6}\textbf{98.45$\pm$0.17} & \cellcolor{gray!6}{98.39$\pm$0.28}\\
F1 & 77.21$\pm$2.24 & 82.84$\pm$2.04 & 87.86$\pm$2.02 & \textbf{90.03$\pm$1.05} & 89.76$\pm$1.77\\
\cellcolor{gray!6}{coverage} & \cellcolor{gray!6}\textbf{95.56$\pm$1.67} & \cellcolor{gray!6}{95.41$\pm$1.49} & \cellcolor{gray!6}{94.99$\pm$1.33} & \cellcolor{gray!6}{94.91$\pm$1.77} & \cellcolor{gray!6}{94.01$\pm$1.36}\\
Jaccard & 0.46$\pm$0.05 & 0.73$\pm$0.03 & 0.80$\pm$0.03 & \textbf{0.88$\pm$0.02} & 0.84$\pm$0.02\\
\cellcolor{gray!6}{setsize*} & \cellcolor{gray!6}{2.64$\pm$0.31} & \cellcolor{gray!6}{1.58$\pm$0.12} & \cellcolor{gray!6}{1.34$\pm$0.10} & \cellcolor{gray!6}\textbf{1.17$\pm$0.07} & \cellcolor{gray!6}{1.22$\pm$0.06}\\
pctempty & 0.00$\pm$0.00 & 0.00$\pm$0.00 & 0.25$\pm$0.46 & 0.12$\pm$0.37 & 0.00$\pm$0.00\\
\cellcolor{gray!6}{MCriterion*} & \cellcolor{gray!6}{0.82$\pm$0.06} & \cellcolor{gray!6}{0.44$\pm$0.05} & \cellcolor{gray!6}{0.30$\pm$0.07} & \cellcolor{gray!6}\textbf{0.15$\pm$0.04} & \cellcolor{gray!6}{0.20$\pm$0.05}\\
FCriterion* & 0.26$\pm$0.04 & 0.12$\pm$0.02 & 0.08$\pm$0.01 & \textbf{0.06$\pm$0.01} & 0.21$\pm$0.02\\
\cellcolor{gray!6}{OM*} & \cellcolor{gray!6}{0.82$\pm$0.06} & \cellcolor{gray!6}{0.45$\pm$0.04} & \cellcolor{gray!6}{0.33$\pm$0.06} & \cellcolor{gray!6}\textbf{0.19$\pm$0.04} & \cellcolor{gray!6}{0.25$\pm$0.04}\\
OF* & 0.24$\pm$0.04 & 0.09$\pm$0.01 & 0.05$\pm$0.01 & \textbf{0.03$\pm$0.01} & 0.05$\pm$0.01\\
\cellcolor{gray!6}{OU*} & \cellcolor{gray!6}{0.16$\pm$0.02} & \cellcolor{gray!6}{0.07$\pm$0.01} & \cellcolor{gray!6}{0.04$\pm$0.01} & \cellcolor{gray!6}\textbf{0.02$\pm$0.01} & \cellcolor{gray!6}{0.04$\pm$0.01}\\
OE* & 1.68$\pm$0.30 & 0.62$\pm$0.11 & 0.39$\pm$0.10 & \textbf{0.23$\pm$0.06} & 0.28$\pm$0.05\\
\bottomrule
\end{tabular}
\end{table}

\begin{figure}[h!]
\centering
\includegraphics[width=.5\textwidth]{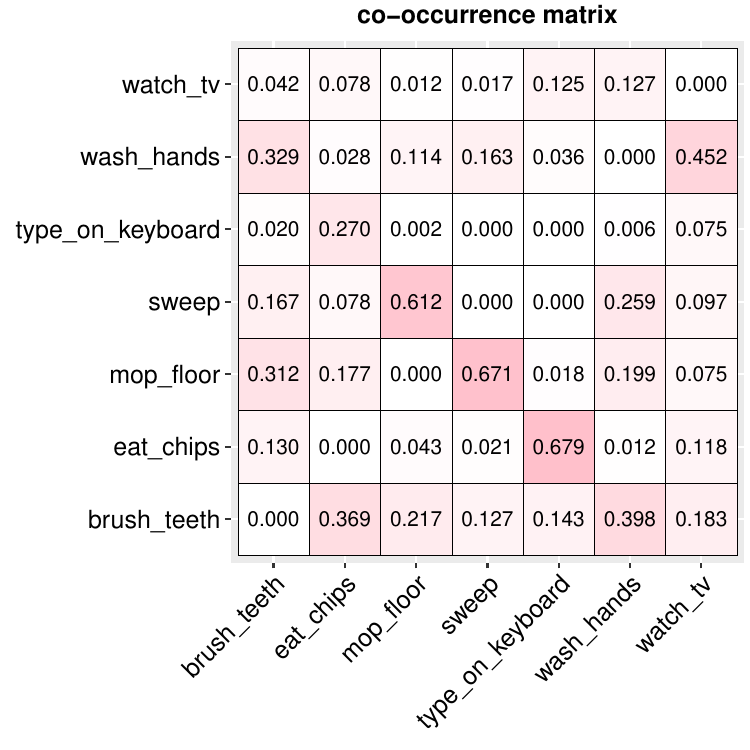}\hfill
\includegraphics[width=.5\textwidth]{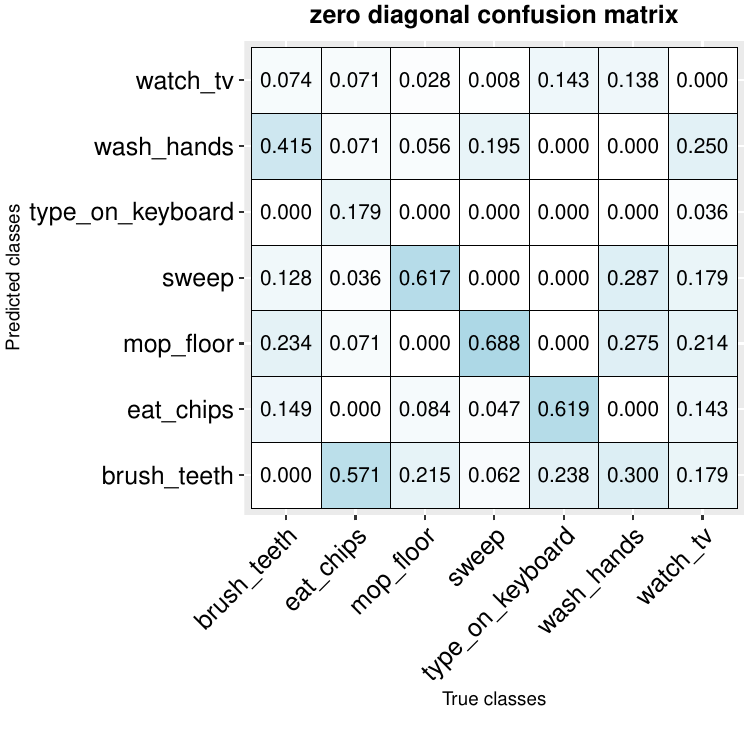}\hfill
\caption{HTAD dataset with MV-S. Co-occurrence matrix (left), zero diagonal confusion matrix (right).}
\label{fig:htad-plots}
\end{figure}

Figure~\ref{fig:htad-plots} shows the \emph{co-occurrence matrix} and the \emph{zero diagonal confusion matrix} for the HTAD dataset using MV-S. These two types of plots were proposed in~\cite{garcia2023conformal}. Those plots are useful for inspecting the results of conformal prediction sets. The co-occurrence matrix depicts the frequencies with which every label occurred in the same prediction set with every other label. The frequency of a class with itself (the diagonal) is set to zero since every class always occurs with itself. The reason is to emphasize the patterns with other labels. The columns are also normalized so they sum up to $1$. The zero-diagonal confusion matrix is similar to the normal confusion matrix but it is column normalized and the diagonal is set to zero so it can be compared with the co-occurrence matrix. In the co-occurrence matrix, it can be seen that \emph{`mop\_floor'} (third column) co-occurred with \emph{`sweep'} in $61.2\%$ of the prediction sets containing \emph{`mop\_floor'}. In the zero-diagonal confusion matrix, it can also be seen that \emph{`mop\_floor'} was confused with \emph{`sweep'} $61.7\%$ of the time. Both plots present similar patterns. That is, if a class $X$ co-occurred with high probability with a class $Y$ then it was also likely that $X$ was confused with $Y$ by the non-conformal model.

\begin{figure}[!h]
\centering
\includegraphics[scale=0.5]{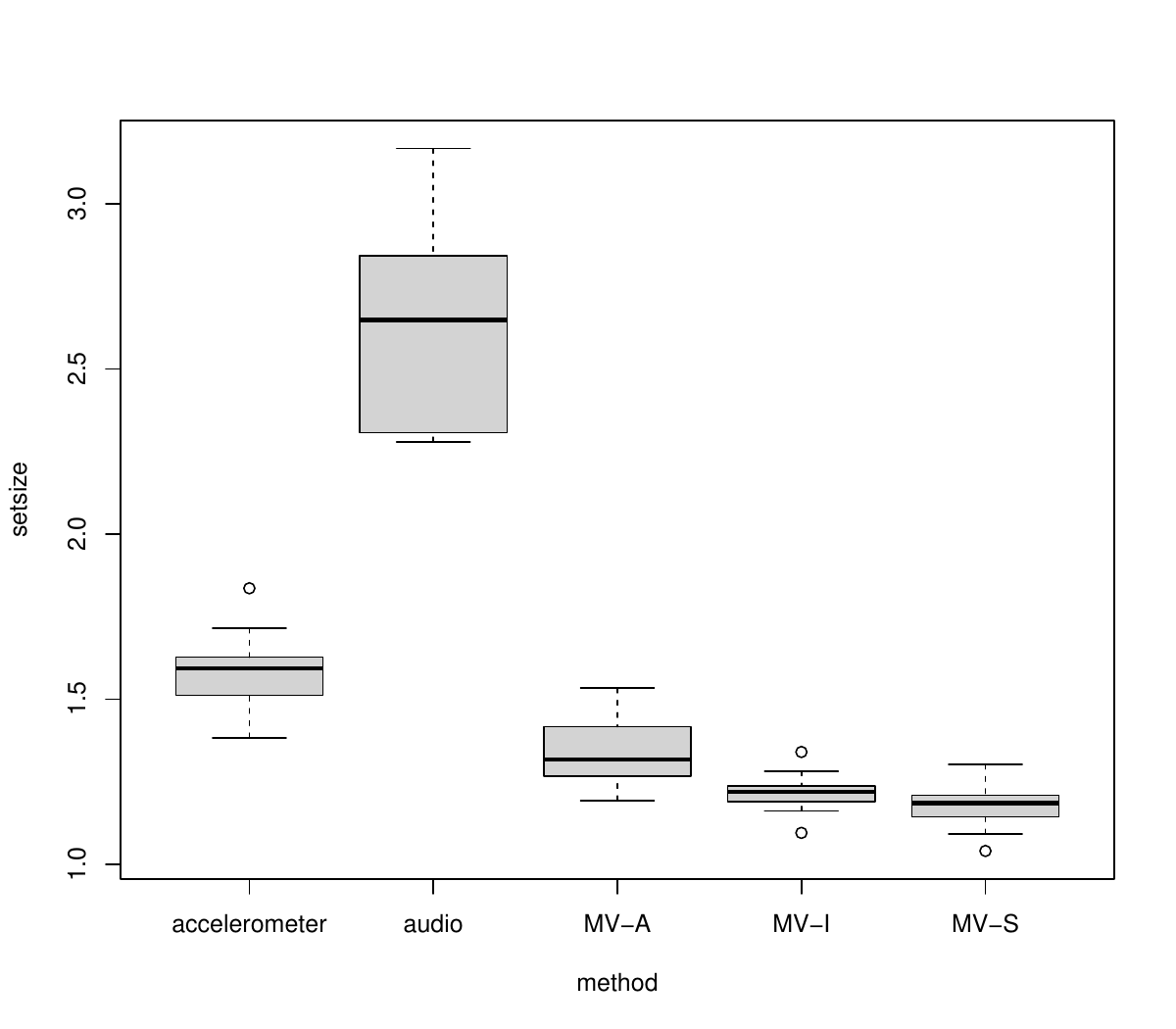}
\caption{Average set-size for the HTAD Dataset. Smaller values are better.}
\label{fig:htad_boxplot_setsize}
\end{figure}

\begin{figure}[!h]
\centering
\includegraphics[scale=0.5]{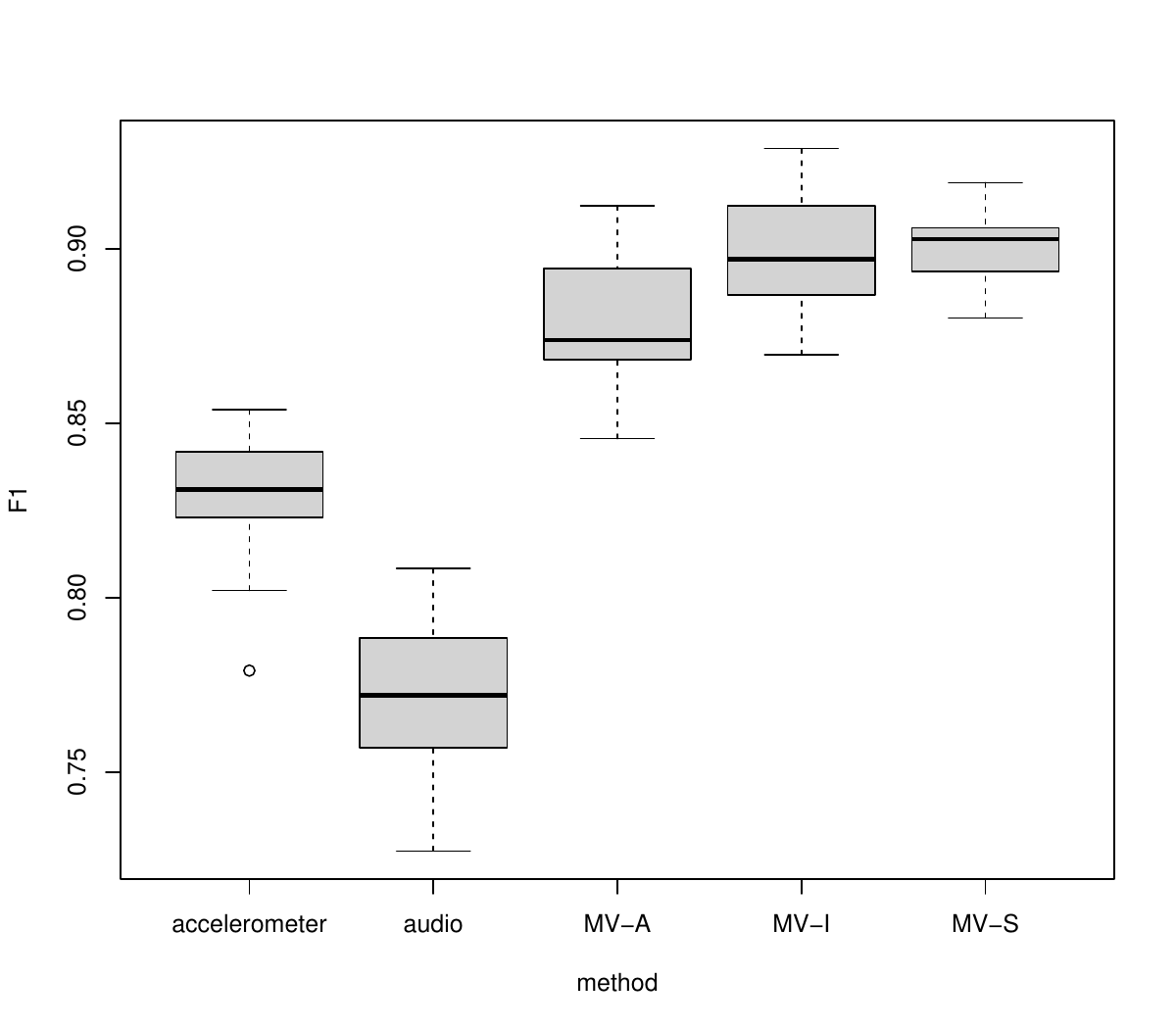}
\caption{Average F1-Score for the HTAD Dataset. Bigger values are better.}
\label{fig:htad_boxplot_f1}
\end{figure}

Figure~\ref{fig:htad_boxplot_setsize} shows box plots of the set size for each of the models. In this case, MV-S and MV-I were the best models (smaller values are preferred). Figure~\ref{fig:htad_boxplot_f1} shows box plots for the F1-Score. Again, MV-S and MV-I were the two best performing models. The worst performing model in both cases was the one using only audio information. It can also be noted that all the multi-view models are far better than the single-view models.

\begin{figure}[!h]
\centering
\includegraphics[scale=0.5]{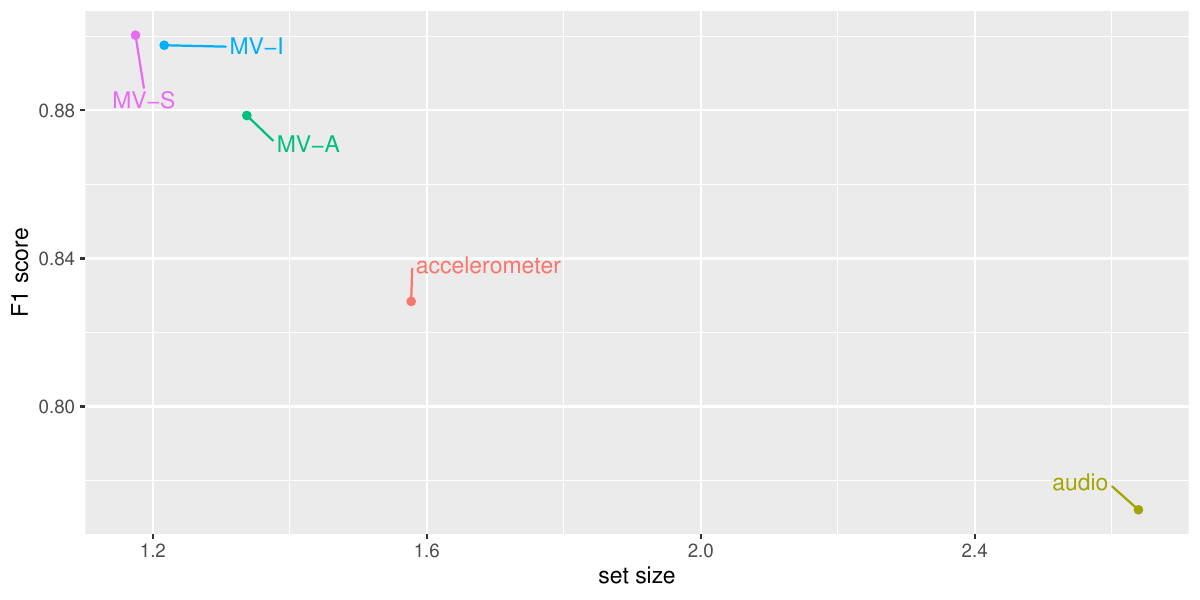}
\caption{Average F1-Score and set-size for the HTAD Dataset.}
\label{fig:htad_scatter}
\end{figure}

In order to compare two measures at a time, we generated a scatter plot of set size vs. F1-Score (Figure~\ref{fig:htad_scatter}). The three multi-view models are clustered in the upper left corner indicating that they performed better in terms of the two measures.

\begin{figure}[h!]
\centering
\includegraphics[width=.5\textwidth]{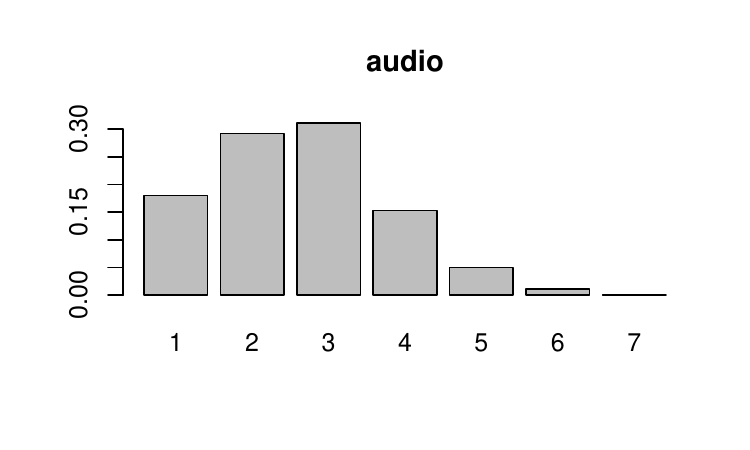}\hfill
\includegraphics[width=.5\textwidth]{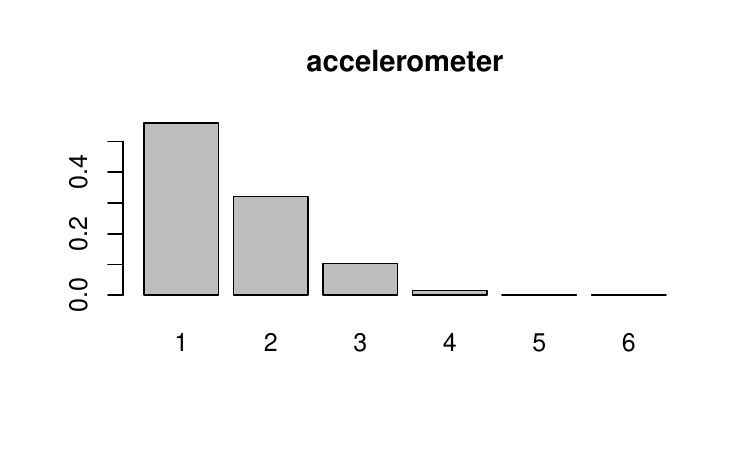}\hfill

\includegraphics[width=.5\textwidth]{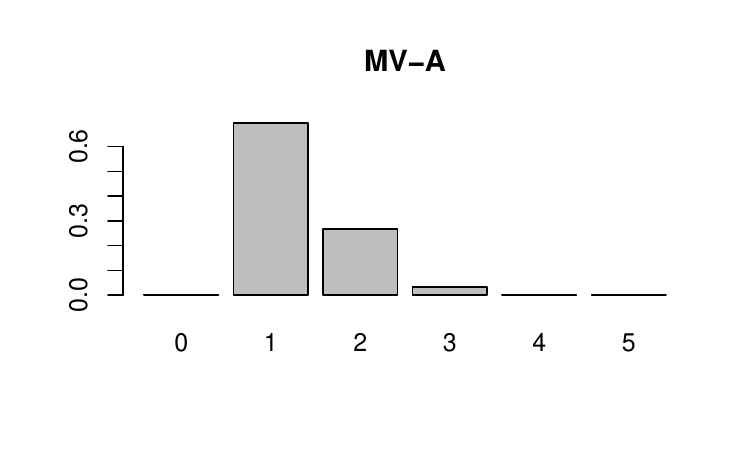}\hfill
\includegraphics[width=.5\textwidth]{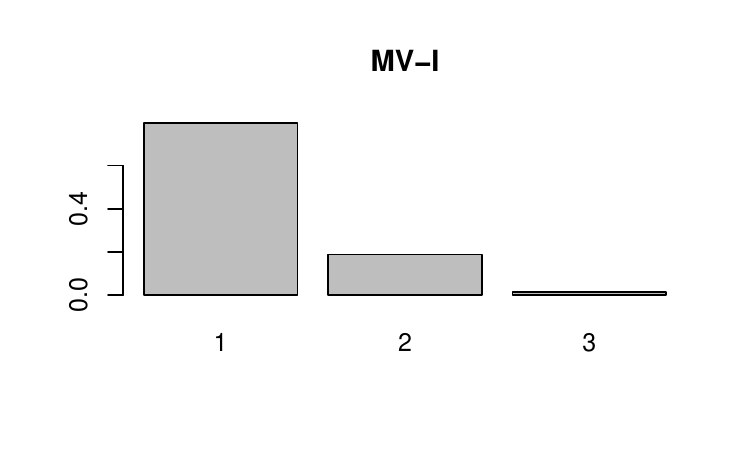}\hfill

\includegraphics[width=.5\textwidth]{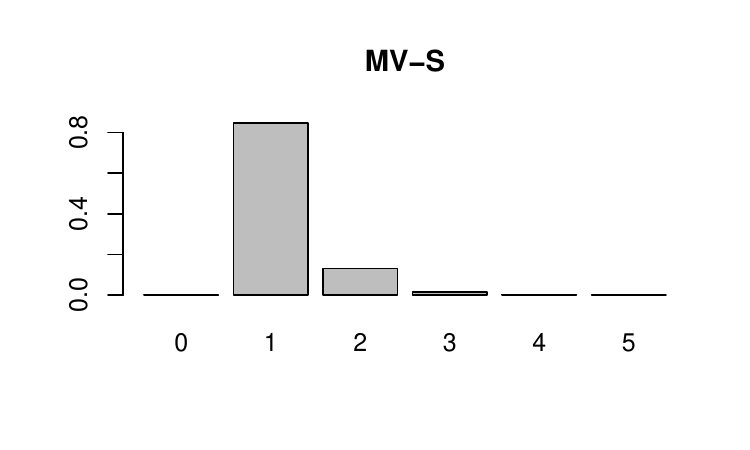}\hfill

\caption{Set size distributions for the HTAD dataset.}
\label{fig:htad_histograms}
\end{figure}

In conformal prediction, sets with small size are preferable since they imply less uncertainty. In order to analyze this, we plotted the set size distributions of all models (Figure~\ref{fig:htad_histograms}). In the case of the model using only audio data, the most frequent set sizes were $2$ and $3$. MV-A and MV-S produced sets ranging from $0$ to $5$ elements. MV-I produced sets in the range of $1-3$ elements which is the expected behavior since MV-I is based on the intersection between the audio and accelerometer prediction sets.

In order to test the performance of the multi-view conformal models with 3 different views, we used the Berkeley-MHAD dataset. Table~\ref{tab:berkeley_results} shows the average results across the $15$ iterations. 

\begin{figure}[h!]
\centering
\includegraphics[width=.5\textwidth]{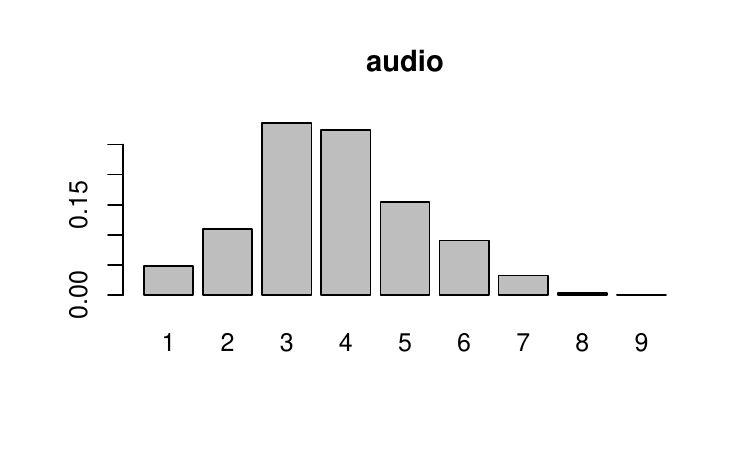}\hfill
\includegraphics[width=.5\textwidth]{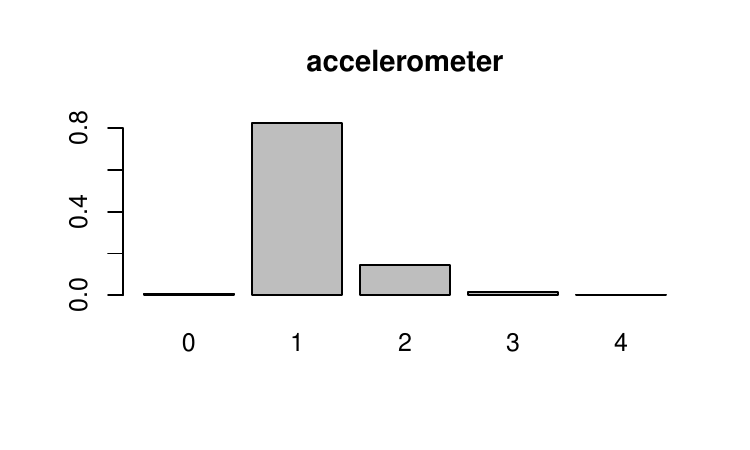}\hfill
\includegraphics[width=.5\textwidth]{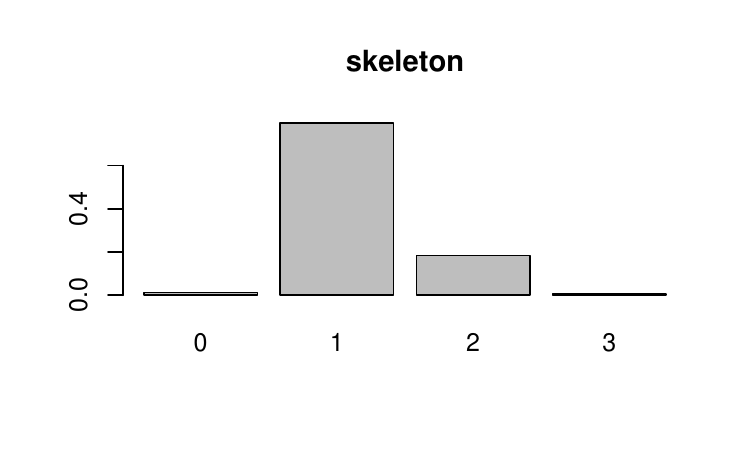}\hfill
\includegraphics[width=.5\textwidth]{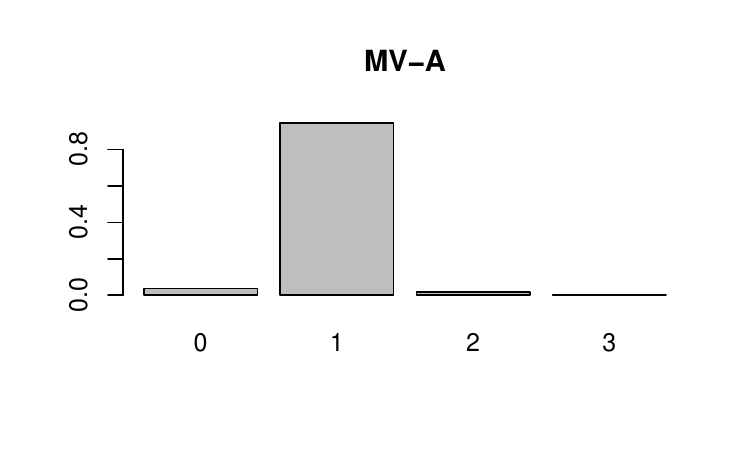}\hfill
\includegraphics[width=.5\textwidth]{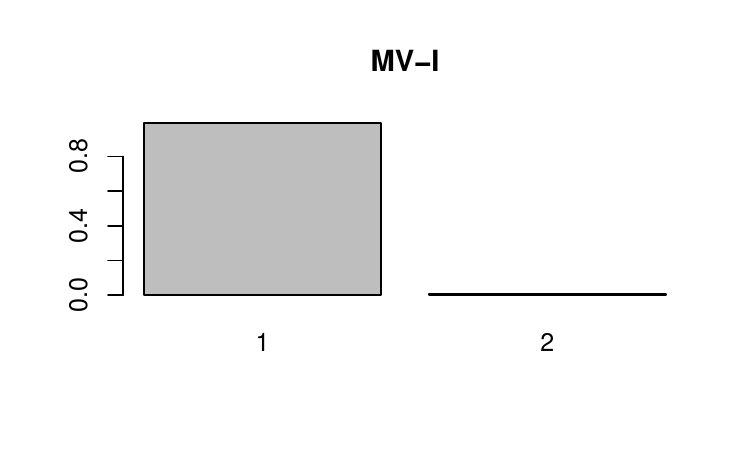}\hfill
\includegraphics[width=.5\textwidth]{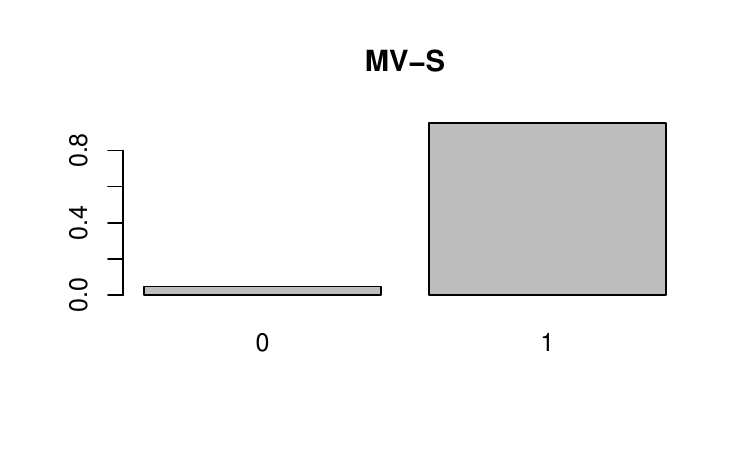}\hfill

\caption{Set size distributions for the Berkeley-MHAD dataset.}
\label{fig:berkeley_histograms}
\end{figure}

Similar to the other dataset, MV-S achieved the best performance in the majority of measures. The M criterion score was $0.0$ since it measures the proportion of prediction sets with more than one element but MV-S only produced sets with one element and some empty sets (Figure~\ref{fig:berkeley_histograms}). The OU criterion was $0.0$ because of the two decimal rounding. The precise score was $0.001889$ but still very small. This is because this measure is based on the largest p-value of the false classes and the proportion of false classes was very low. The OF was also $0.0$ because of the rounding. The more precise value was $0.001889$. Similarly, the OF and OU measures for MV-A were $0.003208$ and $0.003186$, respectively.

\begin{table}[h!]
\small
\caption{Berkeley-MHAD average results. * Smaller values are preferred.}
\label{tab:berkeley_results}
\centering
\begin{tabular}[t]{lllllll}
\toprule
  & \textbf{audio} & \textbf{acc.} & \textbf{skeleton} & \textbf{MV-A} & \textbf{MV-S} & \textbf{MV-I}\\
\midrule
\cellcolor{gray!6}{accuracy} & \cellcolor{gray!6}{58.75$\pm$3.92} & \cellcolor{gray!6}{91.68$\pm$2.07} & \cellcolor{gray!6}{91.52$\pm$2.16} & \cellcolor{gray!6}{96.65$\pm$1.58} & \cellcolor{gray!6}\textbf{97.54$\pm$1.31} & \cellcolor{gray!6}{96.36$\pm$1.55}\\
sensitivity & 58.75$\pm$3.92 & 91.68$\pm$2.07 & 91.52$\pm$2.16 & 96.65$\pm$1.58 & \textbf{97.54$\pm$1.31} & 96.36$\pm$1.55\\
\cellcolor{gray!6}{specificity} & \cellcolor{gray!6}{95.87$\pm$0.39} & \cellcolor{gray!6}{99.17$\pm$0.21} & \cellcolor{gray!6}{99.15$\pm$0.22} & \cellcolor{gray!6}{99.66$\pm$0.16} & \cellcolor{gray!6}\textbf{99.75$\pm$0.13} & \cellcolor{gray!6}{99.64$\pm$0.16}\\
F1 & 58.14$\pm$3.92 & 91.57$\pm$2.09 & 91.43$\pm$2.19 & 96.64$\pm$1.59 & \textbf{97.52$\pm$1.31} & 96.35$\pm$1.56\\
\cellcolor{gray!6}{coverage} & \cellcolor{gray!6}{95.88$\pm$2.79} & \cellcolor{gray!6}{95.60$\pm$2.46} & \cellcolor{gray!6}{95.15$\pm$2.06} & \cellcolor{gray!6}{94.87$\pm$2.97} & \cellcolor{gray!6}{94.34$\pm$2.61} & \cellcolor{gray!6}\textbf{96.73$\pm$1.43}\\
Jaccard & 0.31$\pm$0.05 & 0.87$\pm$0.07 & 0.86$\pm$0.03 & 0.94$\pm$0.02 & 0.94$\pm$0.03 & \textbf{0.96$\pm$0.02}\\
\cellcolor{gray!6}{setsize*} & \cellcolor{gray!6}{3.80$\pm$0.65} & \cellcolor{gray!6}{1.18$\pm$0.22} & \cellcolor{gray!6}{1.18$\pm$0.09} & \cellcolor{gray!6}{0.98$\pm$0.04} & \cellcolor{gray!6}\textbf{0.95$\pm$0.03} & \cellcolor{gray!6}{1.01$\pm$0.01}\\
pctempty & 0.00$\pm$0.00 & 0.89$\pm$0.97 & 1.33$\pm$1.00 & 3.76$\pm$2.81 & 4.81$\pm$2.73 & 0.00$\pm$0.00\\
\cellcolor{gray!6}{MCriterion*} & \cellcolor{gray!6}{0.95$\pm$0.03} & \cellcolor{gray!6}{0.17$\pm$0.14} & \cellcolor{gray!6}{0.19$\pm$0.07} & \cellcolor{gray!6}{0.02$\pm$0.02} & \cellcolor{gray!6}\textbf{0.00$\pm$0.00} & \cellcolor{gray!6}{0.01$\pm$0.01}\\
FCriterion* & 0.62$\pm$0.10 & 0.11$\pm$0.03 & 0.09$\pm$0.01 & \textbf{0.07$\pm$0.00} & 0.07$\pm$0.03 & 0.31$\pm$0.04\\
\cellcolor{gray!6}{OM*} & \cellcolor{gray!6}{0.95$\pm$0.02} & \cellcolor{gray!6}{0.20$\pm$0.13} & \cellcolor{gray!6}{0.22$\pm$0.07} & \cellcolor{gray!6}{0.03$\pm$0.02} & \cellcolor{gray!6}\textbf{0.01$\pm$0.01} & \cellcolor{gray!6}{0.04$\pm$0.02}\\
OF* & 0.64$\pm$0.10 & 0.03$\pm$0.01 & 0.03$\pm$0.01 & \textbf{0.00$\pm$0.00} & \textbf{0.00$\pm$0.00} & 0.01$\pm$0.00\\
\cellcolor{gray!6}{OU*} & \cellcolor{gray!6}{0.38$\pm$0.03} & \cellcolor{gray!6}{0.02$\pm$0.01} & \cellcolor{gray!6}{0.03$\pm$0.01} & \cellcolor{gray!6}\textbf{0.00$\pm$0.00} & \cellcolor{gray!6}\textbf{0.00$\pm$0.00} & \cellcolor{gray!6}{0.01$\pm$0.00}\\
OE* & 2.84$\pm$0.63 & 0.23$\pm$0.21 & 0.23$\pm$0.08 & 0.03$\pm$0.02 & \textbf{0.01$\pm$0.01} & 0.04$\pm$0.02\\
\bottomrule
\end{tabular}
\end{table}

\begin{figure}[h!]
\centering
\includegraphics[width=.5\textwidth]{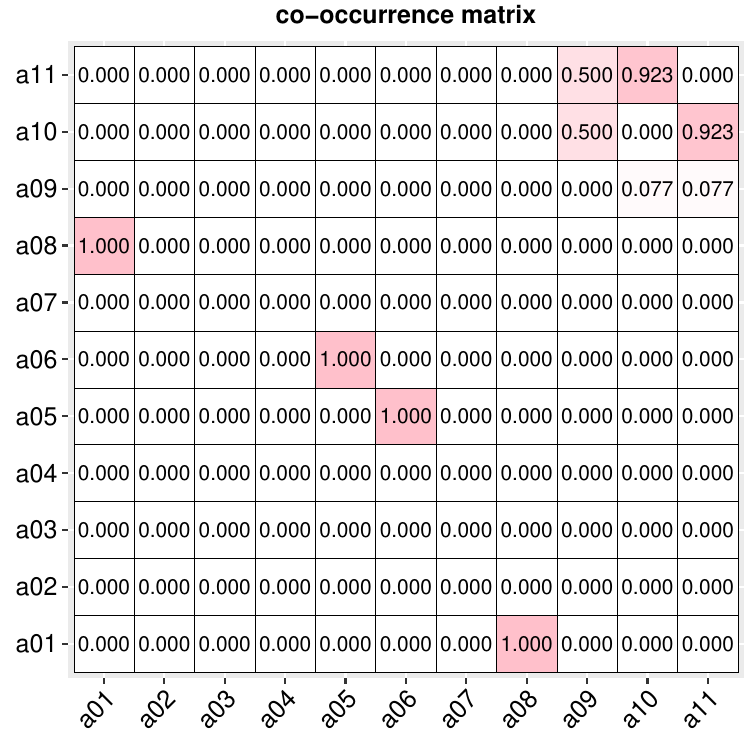}\hfill
\includegraphics[width=.5\textwidth]{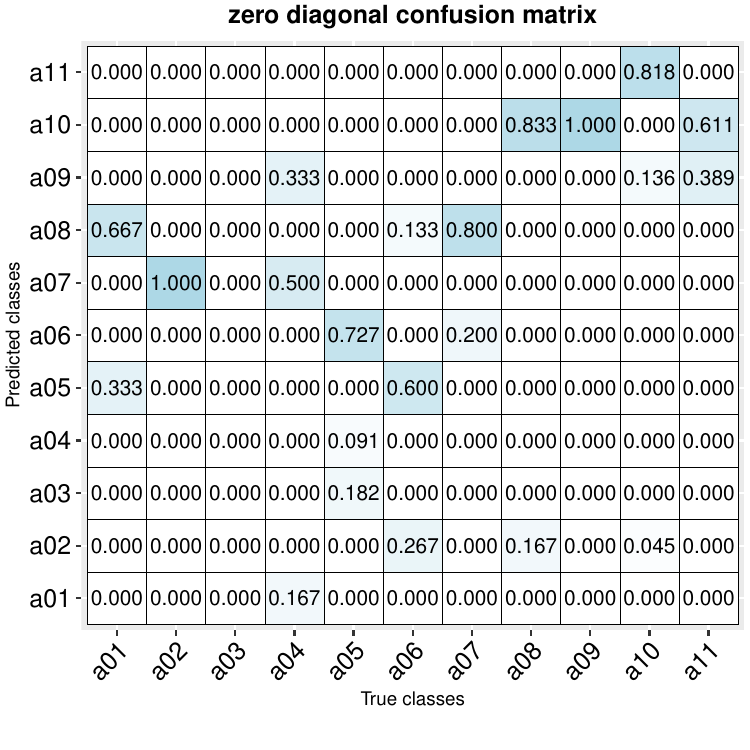}\hfill
\caption{Berkeley-MHAD dataset with MV-I. Co-occurrence matrix (left), zero diagonal confusion matrix (right).}
\label{fig:berkeley-plots}
\end{figure}

Figure~\ref{fig:berkeley-plots} shows the corresponding prediction visualizations for the MV-I model. Again, similar patterns are present in both plots. For example, \emph{`a10 (sit down)'} frequently co-occurred with \emph{`a11 (stand up)'} and those two classes were also frequently confused. The plots for the rest of the models are included in~\ref{sec:appendix}.

\begin{figure}[!h]
\centering
\includegraphics[scale=0.5]{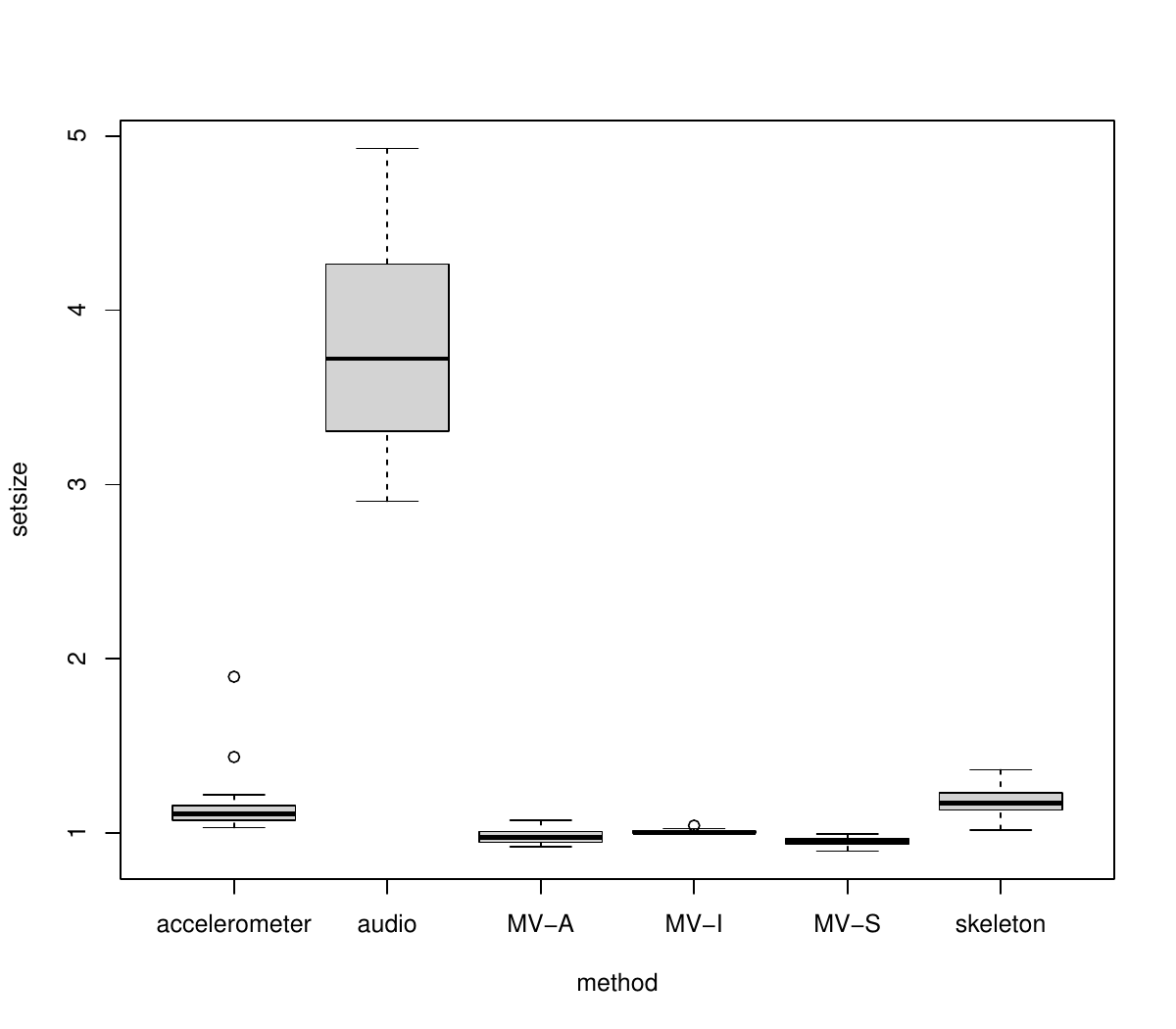}
\caption{Average set-size for the Berkeley-MHAD Dataset. Smaller values are better.}
\label{fig:berkeley_boxplot_setsize}
\end{figure}

\begin{figure}[!h]
\centering
\includegraphics[scale=0.5]{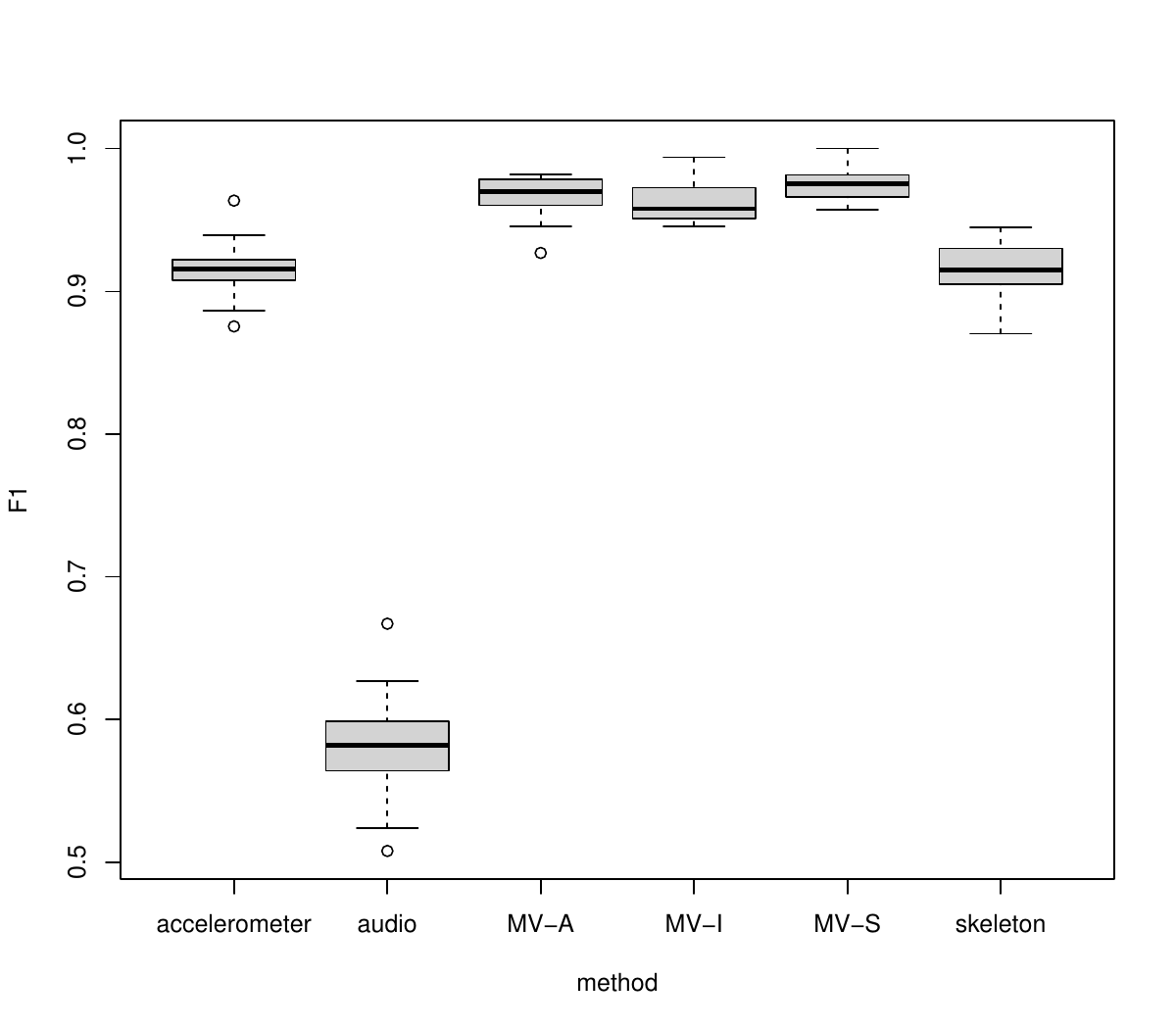}
\caption{Average F1-Score for the Berkeley-MHAD Dataset. Bigger values are better.}
\label{fig:berkeley_boxplot_f1}
\end{figure}

Figures~\ref{fig:berkeley_boxplot_setsize} and~\ref{fig:berkeley_boxplot_f1} show box plots of the set size and F1-Score, respectively. The set sizes of the multi-view models were smaller than the single-view models. The F1-Scores of the multi-view models were higher than the single-view models.

\begin{figure}[!h]
\centering
\includegraphics[scale=0.5]{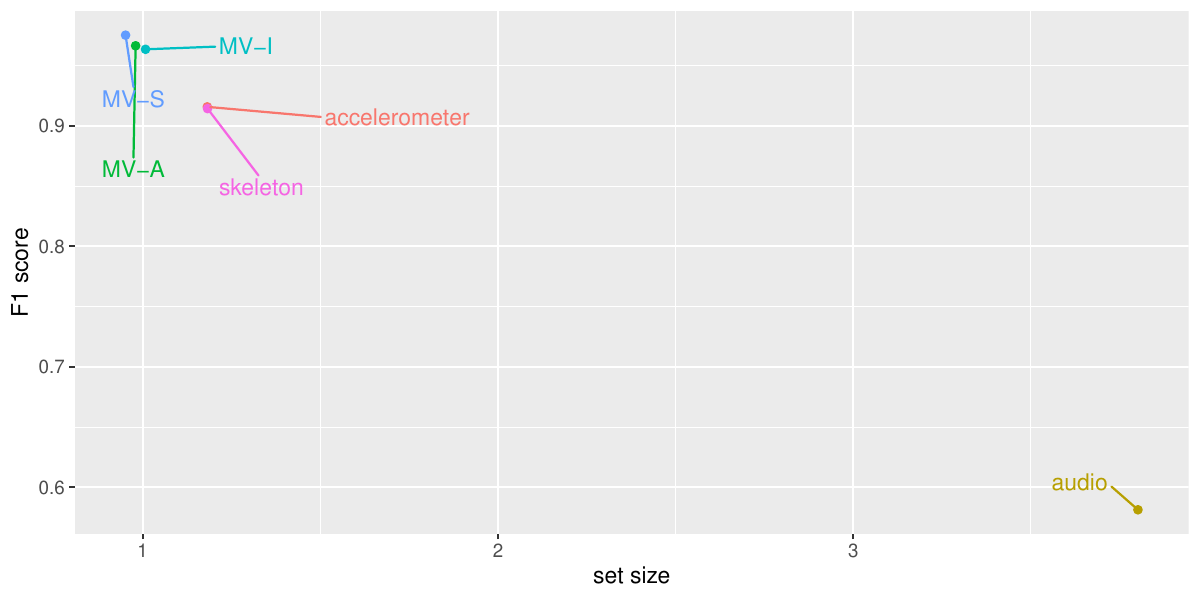}
\caption{Average F1-Score and set-size for the Berkeley-MHAD Dataset.}
\label{fig:berkeley_scatter}
\end{figure}

Based on the set size and F1-Score (Figure~\ref{fig:berkeley_scatter}) one can see that all the multi-view models were better than the single-view models.

The set size distributions (Figure~\ref{fig:berkeley_histograms}) indicate that the model using only audio data produced the sets with higher uncertainty. It even produced some sets with $9$ elements. MV-I and MV-S generated the smallest average set sizes. 

Based on the results, the first hypothesis is not rejected. That is, multi-view models produce prediction sets with less uncertainty than their single-view counterpart models. To further validate this, we conducted a \emph{t-}test to compare the mean set sizes between the multi-view and single-view models. The difference of means was statistically significant ($p << 0.0001$) in both datasets.

The second hypothesis is not rejected and thus, we can conclude that multi-view models perform better than single-view models not only in terms of traditional metrics but also with respect to conformal measures. We confirmed this with a \emph{t-}test for all conformal metrics to test the alternative hypothesis of whether the difference in means is not equal to $0$. In all cases, the difference in means was significant ($p << 0.01$) in both datasets. In summary, the multi-view models performed better than the single-view models in terms of traditional and conformal measures. Among the multi-view models, MV-S performed the best in both datasets. MV-I performed better than MV-A in the HTAD dataset. In the Berkely-MHAD dataset, MV-A performed better than MV-I.

\section{Conclusions}
\label{sec:conclusions}

Despite the substantial improvements in the last years in the accuracy performance of machine learning models for sensor fusion, little has been done to provide confidence estimates for individual predictions. To this extent, in this work we built and tested multi-view conformal models for sensor fusion that provide confidence estimates. Our models are based on the conformal prediction framework that provides coverage guarantees. Furthermore, we proposed a multi-view semi-conformal model based on the intersection of individual conformal models. We conducted experiments with two datasets that have heterogeneous sensors. Our results showed that the multi-view models performed better than the single-view models in both, traditional and conformal measures. On average, the multi-view models produced smaller prediction sets and higher F1-Scores. The MV-S model obtained the best results in both datasets in the majority of measures. In this work we used Random Forest as the underlying model for all of our experiments. However, different models may be more suitable for each type of view. Thus, it is left as future work to test heterogeneous underlying models depending on the type of view. Furthermore, the selection of the best underlying models can be framed as an optimization problem. Even the selection of sensors can be decided based on some optimization technique such as Bayesian optimization~\cite{garnett2010bayesian}. For our experiments, we tested our models with $2$ and $3$ views. Analyzing the conformal behavior as more views are added is also worth exploring.

\section*{CRediT authorship contribution statement}

\textbf{E.G.C.:} Conceptualization, Methodology, Software, Validation, Formal analysis, Investigation, Writing - Original Draft, Visualization.

\section*{Declaration of competing interest}

The authors declare no competing interests.

\section*{Data availability}

All datasets used in this work are publicly available. See Section~\ref{sec:datasets} for references on where to download them. The code to reproduce the results is available here: \url{https://github.com/enriquegit/multiview-conformal-prediction-paper}.

\appendix

\section{}
\label{sec:appendix}

This section includes the remaining plots for the different types of models and datasets.

\begin{figure}[h!]
\centering
\includegraphics[width=.5\textwidth]{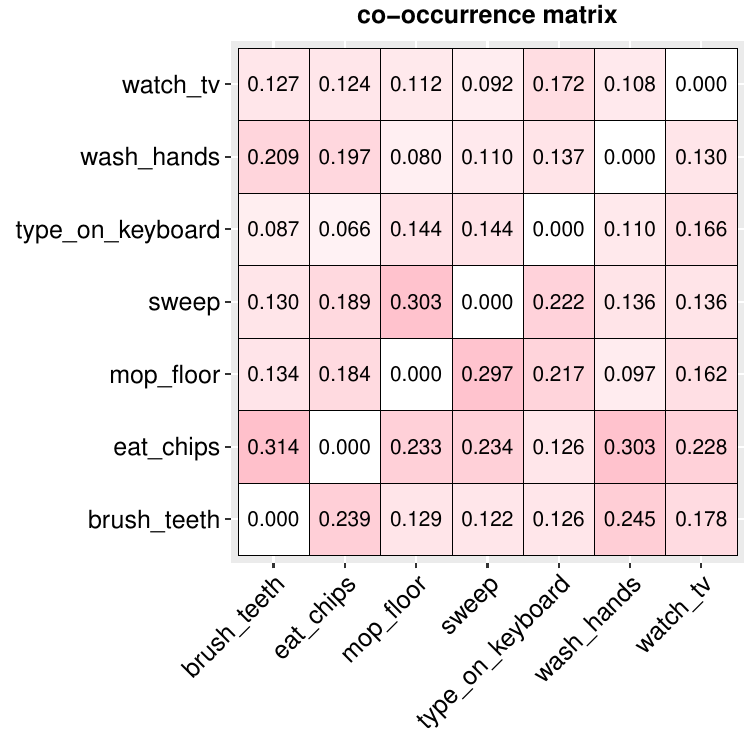}\hfill
\includegraphics[width=.5\textwidth]{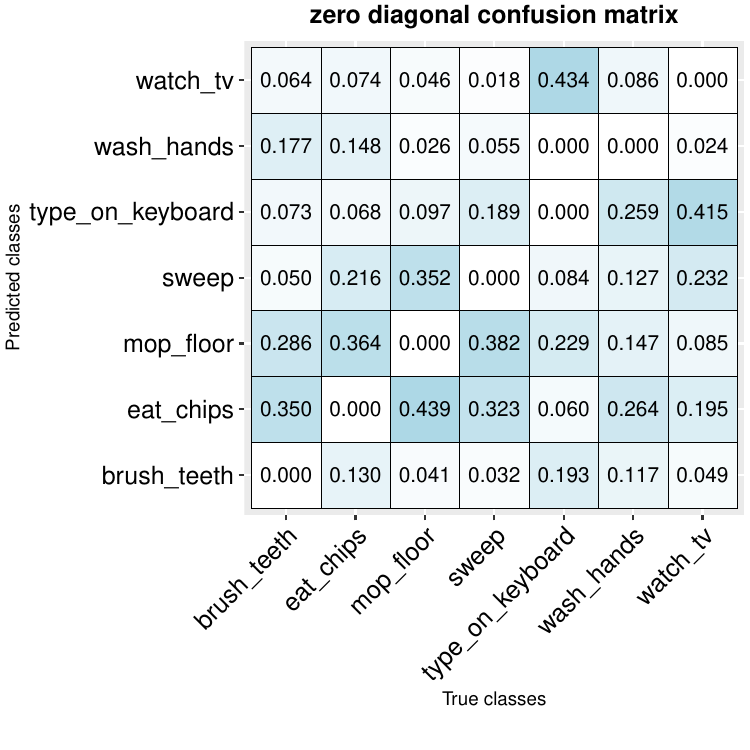}\hfill
\caption{HTAD dataset with audio model. Co-occurrence matrix (left), zero diagonal confusion matrix (right).}
\label{fig:htad-plots-audio}
\end{figure}

\begin{figure}[h!]
\centering
\includegraphics[width=.5\textwidth]{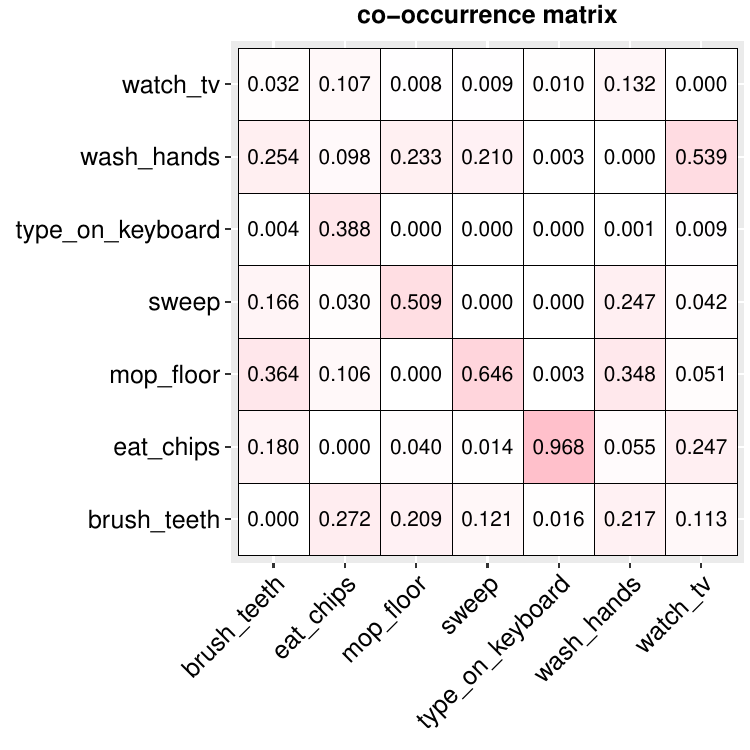}\hfill
\includegraphics[width=.5\textwidth]{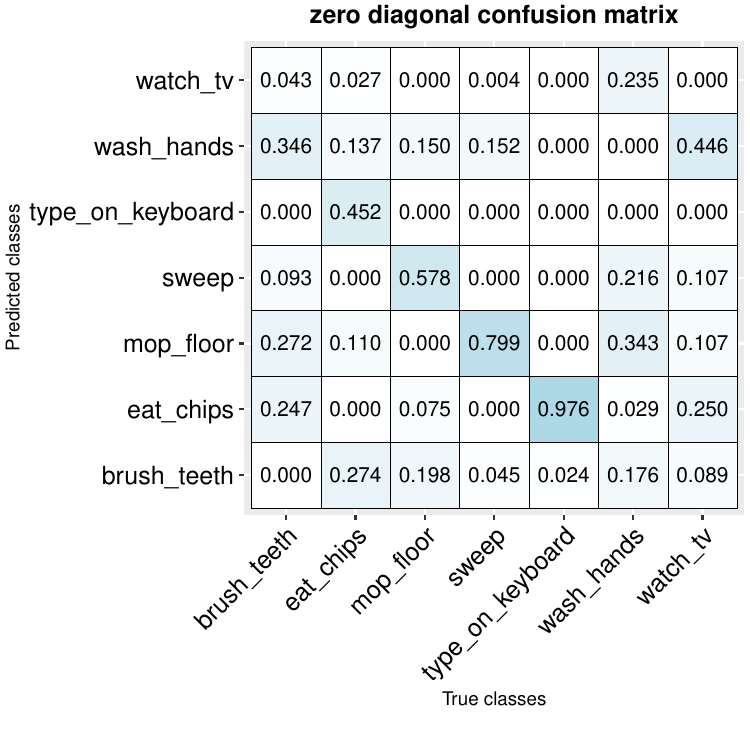}\hfill
\caption{HTAD dataset with accelerometer model. Co-occurrence matrix (left), zero diagonal confusion matrix (right).}
\label{fig:htad-plots-acc}
\end{figure}

\begin{figure}[h!]
\centering
\includegraphics[width=.5\textwidth]{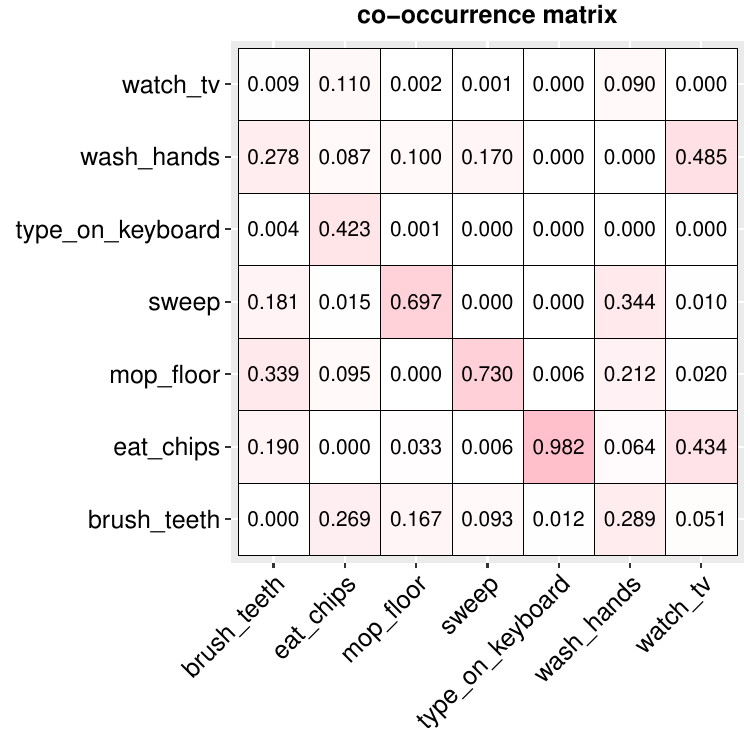}\hfill
\includegraphics[width=.5\textwidth]{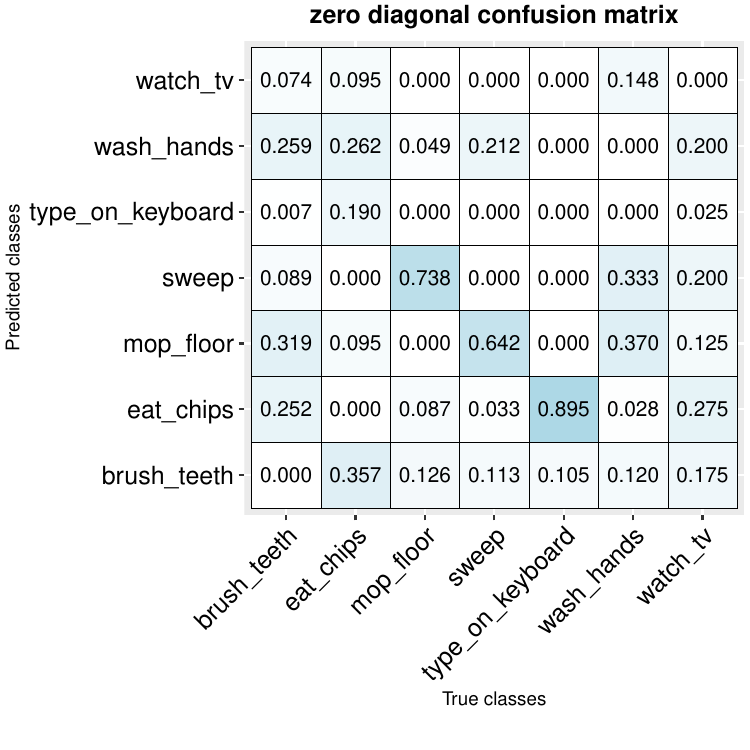}\hfill
\caption{HTAD dataset with MV-A model. Co-occurrence matrix (left), zero diagonal confusion matrix (right).}
\label{fig:htad-plots-mva}
\end{figure}

\begin{figure}[h!]
\centering
\includegraphics[width=.5\textwidth]{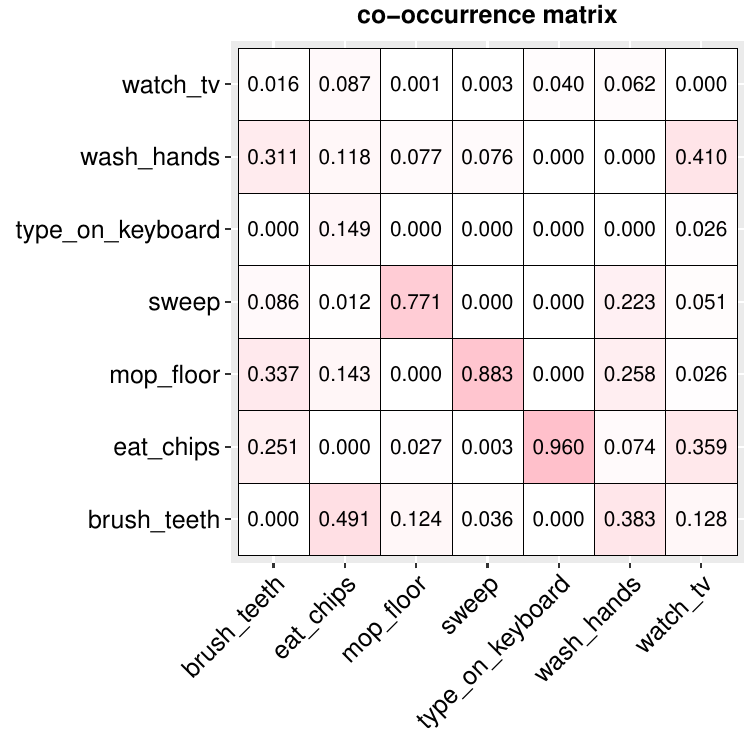}\hfill
\includegraphics[width=.5\textwidth]{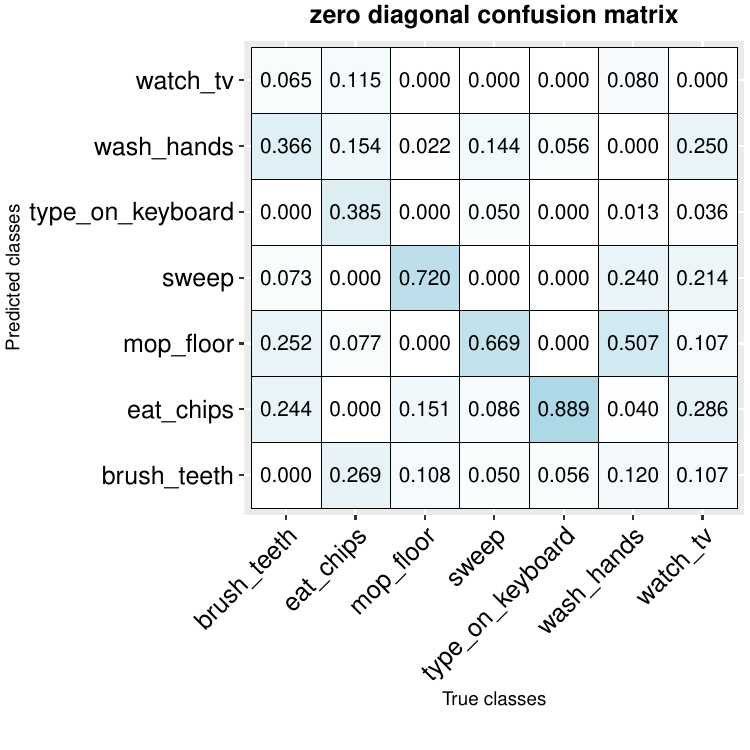}\hfill
\caption{HTAD dataset with MV-I model. Co-occurrence matrix (left), zero diagonal confusion matrix (right).}
\label{fig:htad-plots-mvi}
\end{figure}

\begin{figure}[h!]
\centering
\includegraphics[width=.5\textwidth]{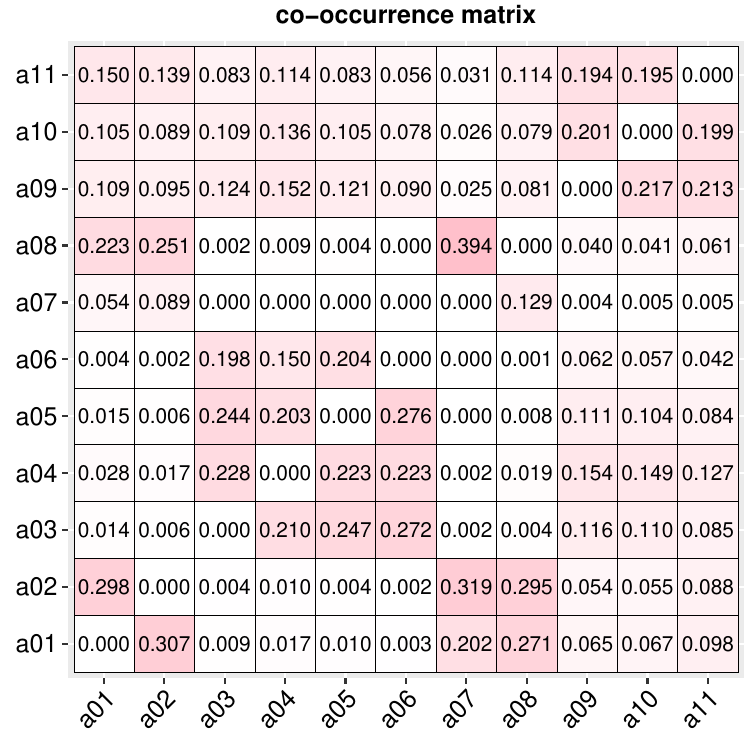}\hfill
\includegraphics[width=.5\textwidth]{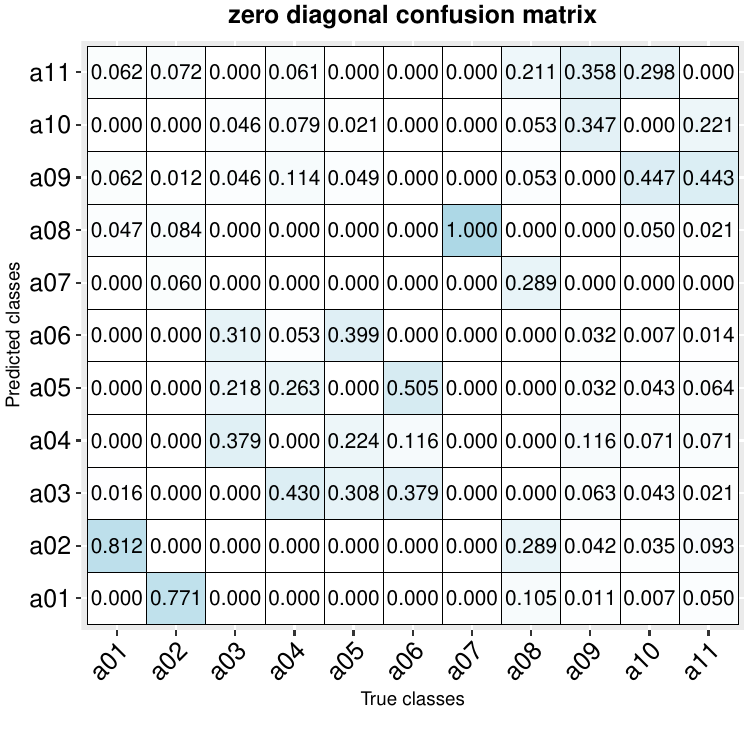}\hfill
\caption{Berkeley-MHAD dataset with audio model. Co-occurrence matrix (left), zero diagonal confusion matrix (right).}
\label{fig:berkeley-plots-audio}
\end{figure}

\begin{figure}[h!]
\centering
\includegraphics[width=.5\textwidth]{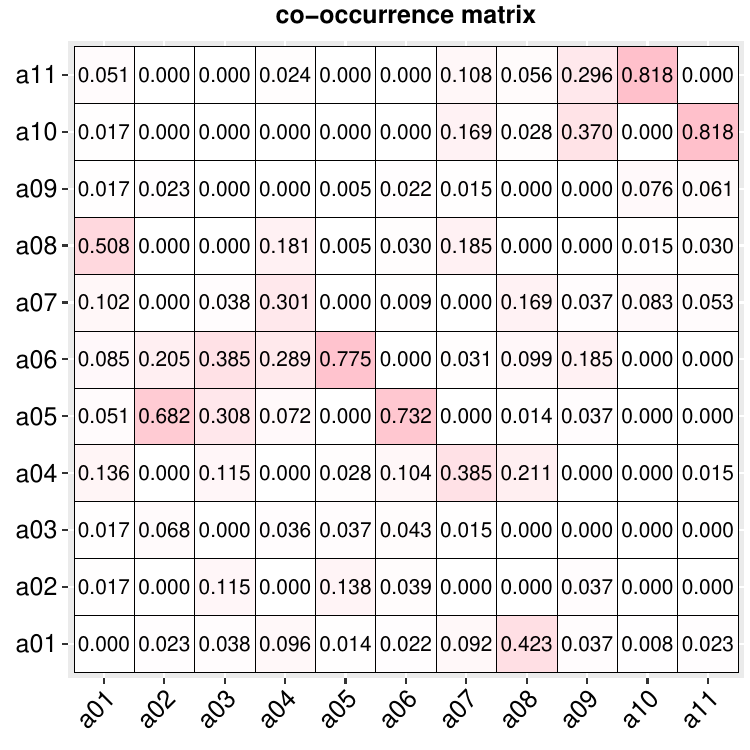}\hfill
\includegraphics[width=.5\textwidth]{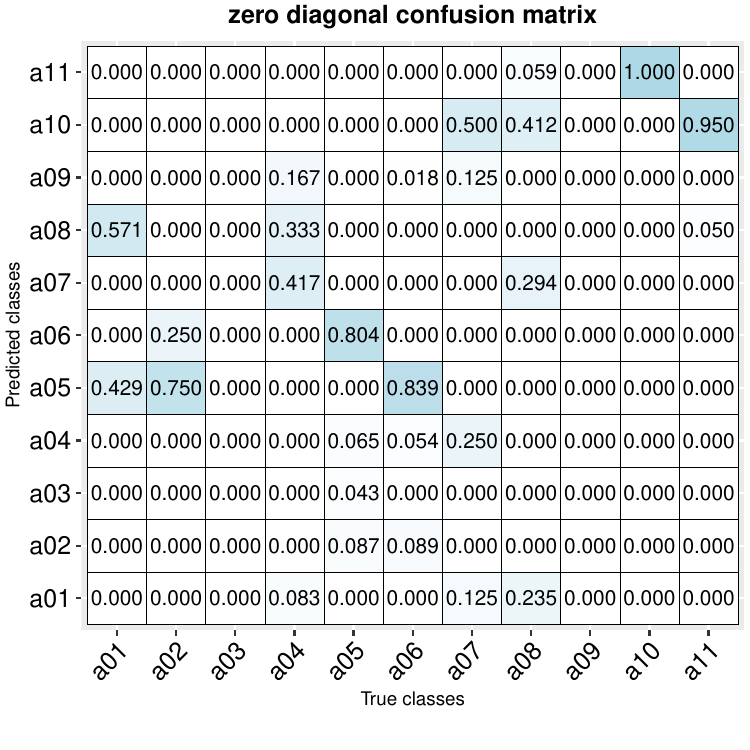}\hfill
\caption{Berkeley-MHAD dataset with accelerometer model. Co-occurrence matrix (left), zero diagonal confusion matrix (right).}
\label{fig:berkeley-plots-acc}
\end{figure}

\begin{figure}[h!]
\centering
\includegraphics[width=.5\textwidth]{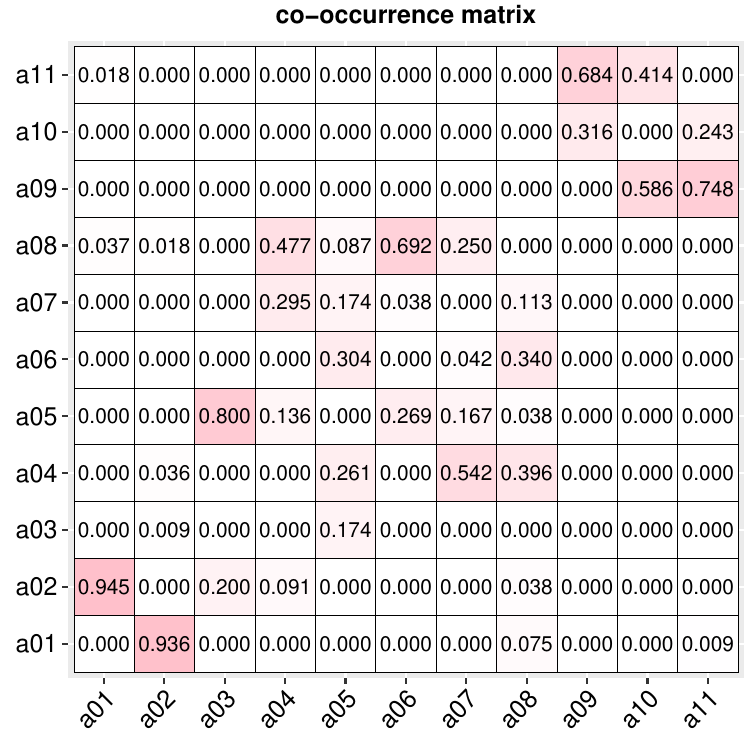}\hfill
\includegraphics[width=.5\textwidth]{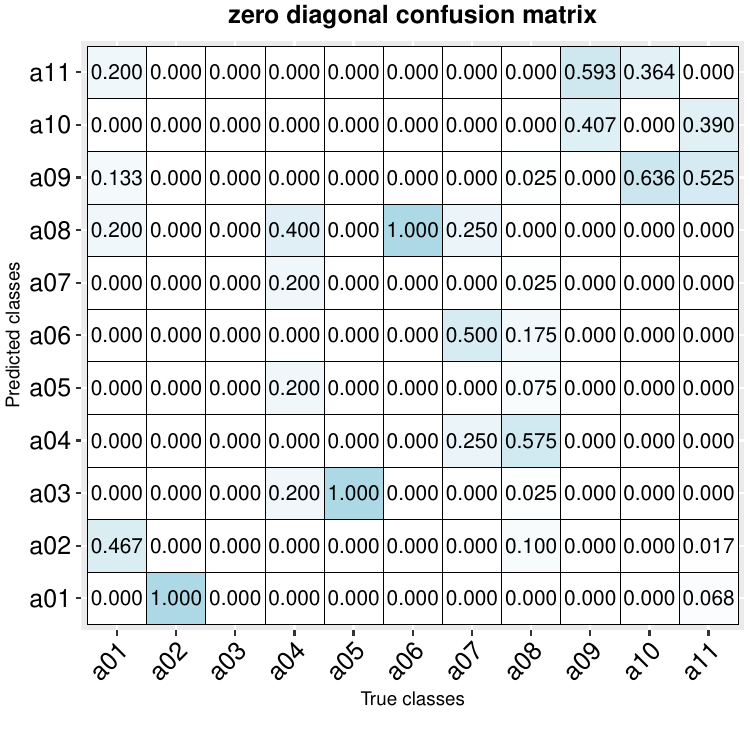}\hfill
\caption{Berkeley-MHAD dataset with skeleton model. Co-occurrence matrix (left), zero diagonal confusion matrix (right).}
\label{fig:berkeley-plots-skel}
\end{figure}

\begin{figure}[h!]
\centering
\includegraphics[width=.5\textwidth]{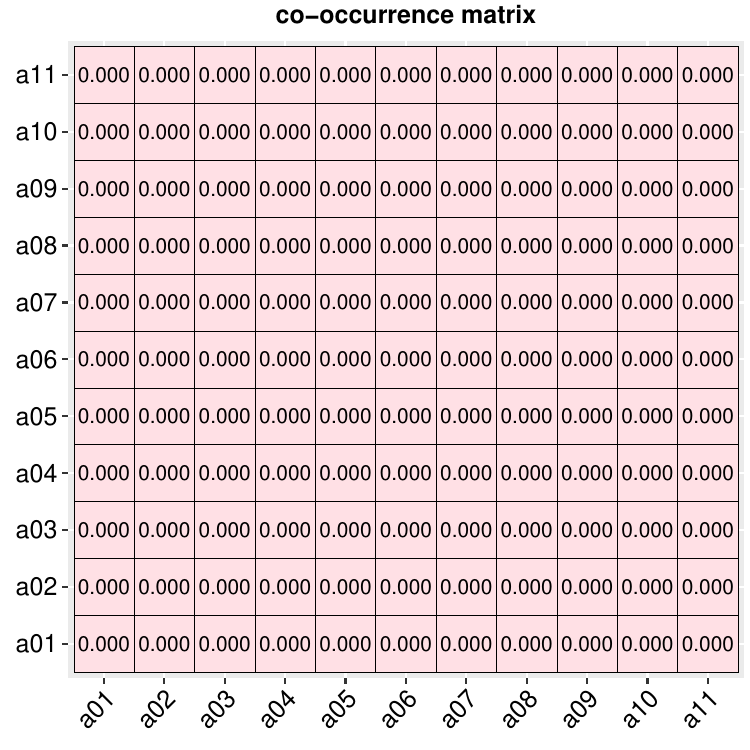}\hfill
\includegraphics[width=.5\textwidth]{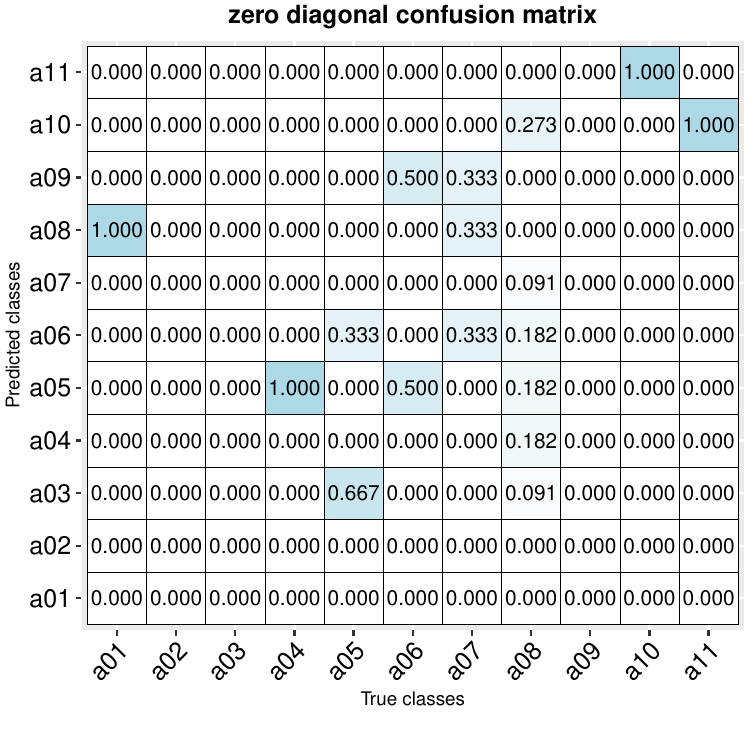}\hfill
\caption{Berkeley-MHAD dataset with MV-S model. Co-occurrence matrix (left), zero diagonal confusion matrix (right).}
\label{fig:berkeley-plots-mvs}
\end{figure}

\begin{figure}[h!]
\centering
\includegraphics[width=.5\textwidth]{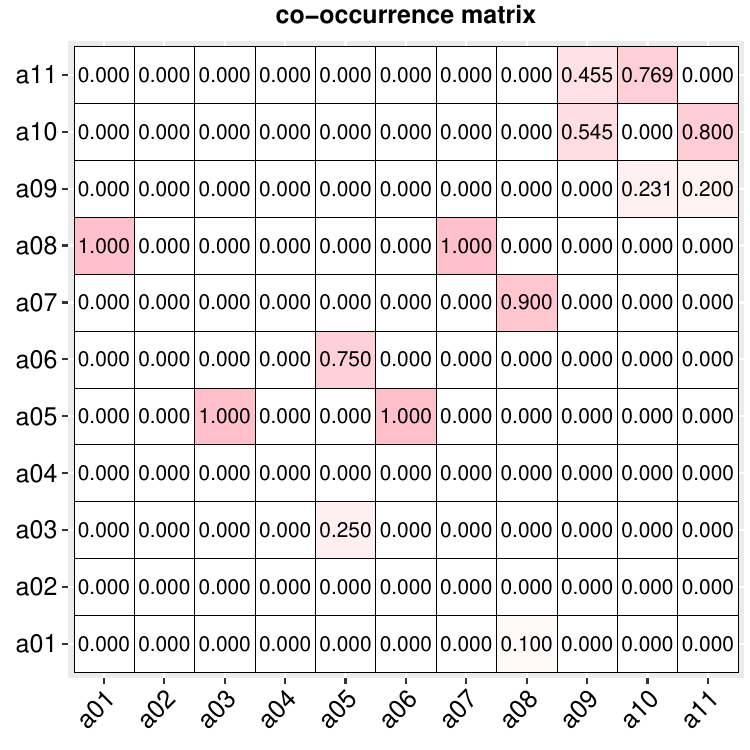}\hfill
\includegraphics[width=.5\textwidth]{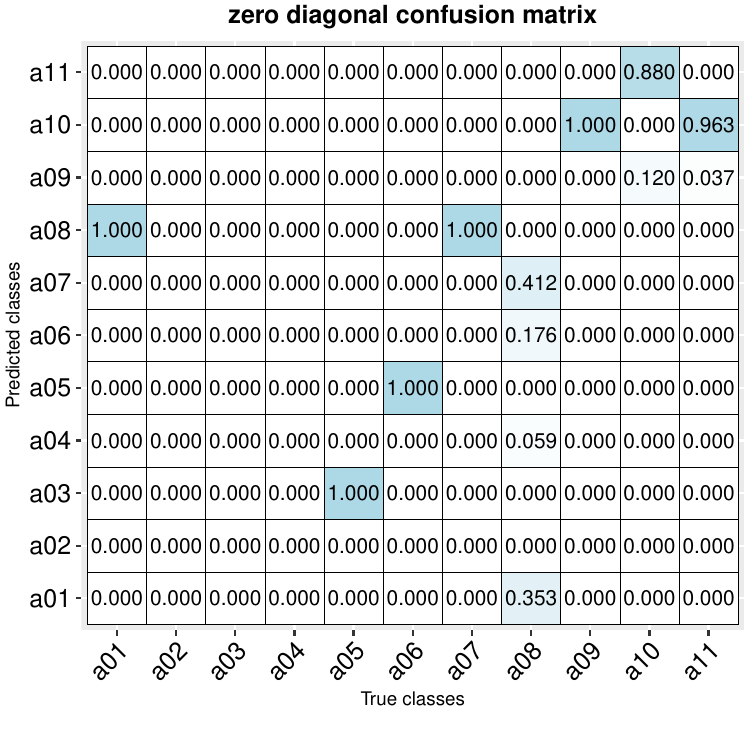}\hfill
\caption{Berkeley-MHAD dataset with MV-A model. Co-occurrence matrix (left), zero diagonal confusion matrix (right).}
\label{fig:berkeley-plots-mva}
\end{figure}

\FloatBarrier
\bibliography{main}





\end{document}